\documentclass[10pt,twocolumn,letterpaper]{article}

\usepackage{iccv}
\usepackage{times}
\usepackage{epsfig}
\usepackage{graphicx}
\usepackage{amsmath}
\usepackage{amssymb}

\usepackage[switch]{lineno}

\usepackage{color}
\usepackage{rotating}
\usepackage{multirow}
\usepackage[lofdepth,lotdepth,caption=false]{subfig}
\definecolor{Gray}{gray}{0.8}
\usepackage{hhline}
\usepackage{colortbl}

\newcommand{\reviewChange}[1]{\textcolor{black}{#1}}

\newcommand{\pointSourceIMG}{x}
\newcommand{\pointTargetIMG}{x'}
\newcommand{\pointSourcePCL}{X}
\newcommand{\pointTargetPCL}{X'}

\newcommand{\keySiftSourceIMG}{\pointSourceIMG_{2d}}
\newcommand{\keySiftTargetIMG}{\pointTargetIMG_{2d}}
\newcommand{\keySiftSourcePCL}{\pointSourcePCL_{2d}}
\newcommand{\keySiftTargetPCL}{\pointTargetPCL_{2d}}

\newcommand{\keyIssSourcePCL}{\pointSourcePCL_{3d}}
\newcommand{\keyIssTargetPCL}{\pointTargetPCL_{3d}}

\newcommand{\keyIcpSourcePCL}{\pointSourcePCL_{icp}}
\newcommand{\keyIcpTargetPCL}{\pointTargetPCL_{icp}}

\newcommand{\RRR}{{\bf R}}
\newcommand{\TTT}{{\bf t}}

\newcommand{\keyContactSourcePCL}{\pointSourcePCL_{hand}}
\newcommand{\keyContactTargetPCL}{\pointTargetPCL_{hand}}

\newcommand{\keyDetectorSourcePCL}{\pointSourcePCL_{det}}
\newcommand{\keyDetectorTargetPCL}{\pointTargetPCL_{det}}

\newcommand{\correspondencesSIFT}{\mathcal{C}_{feat2d}}
\newcommand{\correspondencesISS}{\mathcal{C}_{feat3d}}
\newcommand{\correspondencesCONTACT}{\mathcal{C}_{hand}}
\newcommand{\correspondencesDETECTOR}{\mathcal{C}_{detect}}
\newcommand{\correspondencesICP}{\mathcal{C}_{icp}}

\newcommand{\para}{{\theta}}

\usepackage[pagebackref=true,breaklinks=true,letterpaper=true,colorlinks,bookmarks=false]{hyperref}

\iccvfinalcopy

\def\iccvPaperID{126}

\ificcvfinal\fi

\begin{document}

    \pagenumbering{gobble}

\title{3D Object Reconstruction from Hand-Object Interactions}

\author{
	Dimitrios Tzionas\hspace*{+11mm}\\
	\begin{tabular}[t]{@{}c@{}}
	University of Bonn\\
	Bonn, Germany\\
	{\tt\small \hspace*{22mm} tzionas@iai.uni-bonn.de \hspace*{-22mm}}
	\end{tabular}\nobreak
	\begin{tabular}[t]{@{}c@{}}
	\and
	MPI for Intelligent Systems\\
	T{\"u}bingen, Germany\\
	{}
	\end{tabular}
	\and
	Juergen Gall\\
	University of Bonn,\\
	Bonn, Germany\\
	{\tt\small gall@iai.uni-bonn.de}
}
\maketitle


\begin{abstract}

	Recent advances have enabled 3d object reconstruction approaches using a single off-the-shelf RGB-D camera. 
	Although these approaches are successful for a wide range of object classes, they rely on stable and distinctive geometric or texture features. 
	Many objects like mechanical parts, toys, household or decorative articles, however, are textureless and characterized by minimalistic shapes that are simple and symmetric. 
	Existing in-hand scanning systems and 3d reconstruction techniques fail for such symmetric objects in the absence of highly distinctive features. 
	In this work, we show that extracting 3d hand motion for in-hand scanning effectively facilitates the reconstruction of even featureless and highly symmetric objects and 
	we present an approach that fuses the rich additional information of hands into a 3d reconstruction pipeline, significantly contributing to the state-of-the-art of in-hand scanning.
	
\end{abstract}


\vspace*{-0.5mm}
\section{Introduction}\label{sec:Introduction}
\vspace*{-0.5mm}

	The advent of affordable RGB-D sensors has opened up a whole new range of applications based on the 3d perception of the environment by computers, which includes the creation of a virtual 3d representation of real objects.
	A moving camera can navigate in space observing the real world, while incrementally fusing the acquired frames into a 3d virtual model of it.
	Similarly, 
	a static camera can observe a scene and dynamically reconstruct the observed moving objects.
	This domain has attracted much interest lately in the computer vision, the graphics and the robotics (SLAM) community, as it enables a plethora of other applications, 
	facilitating among others 3d object detection, augmented reality, the internet of things, human-computer-interaction and the interaction of robots with the real world.

	\newcommand{\ImgSquizTeaserH}{-05mm}
	\newcommand{\ImgSquizTeaserV}{-10mm}
	\newcommand{\ImgSizeMultTeaser}{0.20}	
	\newcommand{\ImgSizeMultTeaserr}{0.156}	
	\begin{figure}[t]
	\vspace*{-3mm}
	\captionsetup[subfigure]{labelformat=empty}
	\centering
		\subfloat[subfigure teaser 1][]{	\includegraphics[trim=30mm 15mm 40mm 26mm, clip=true, height=\ImgSizeMultTeaser  \textwidth]{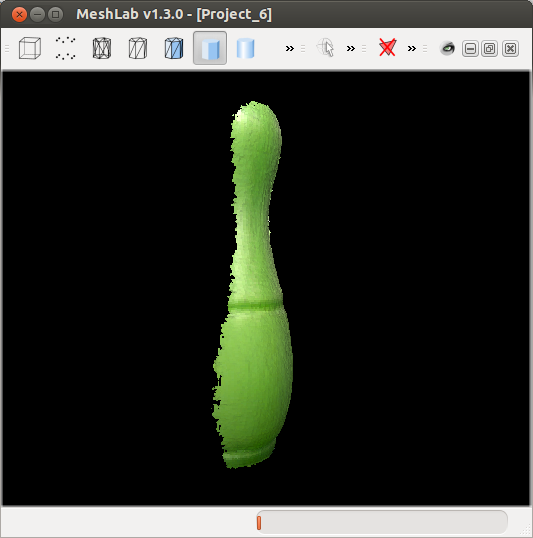}		\label{fig:teaser1}	}		\hspace*{\ImgSquizTeaserH}
		\subfloat[subfigure teaser 2][]{	\includegraphics[trim=40mm 15mm 60mm 30mm, clip=true, height=\ImgSizeMultTeaser  \textwidth]{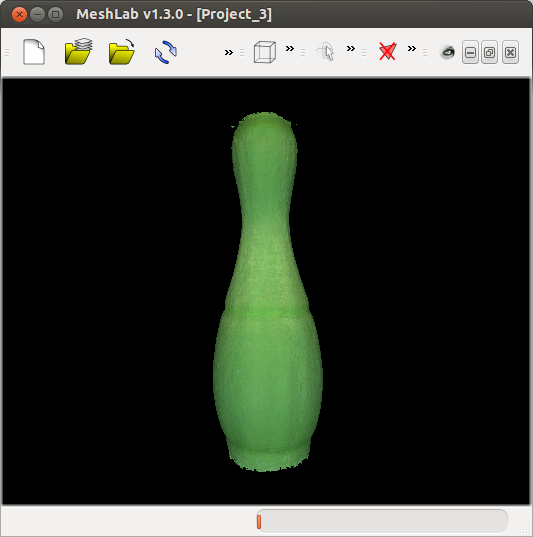}		\label{fig:teaser2}	}		\hspace*{\ImgSquizTeaserH}
		\subfloat[subfigure teaser 3][]{	\includegraphics[trim=20mm 13mm 30mm 30mm, clip=true, height=\ImgSizeMultTeaser  \textwidth]{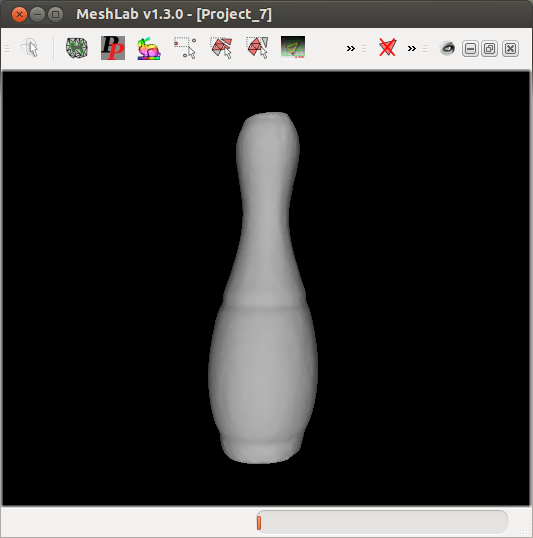}	\label{fig:teaser3}	}		\vspace*{\ImgSquizTeaserV}
		\subfloat[subfigure teaser 4][]{	\includegraphics[trim=30mm 40mm 30mm 50mm, clip=true, width=\ImgSizeMultTeaserr \textwidth]{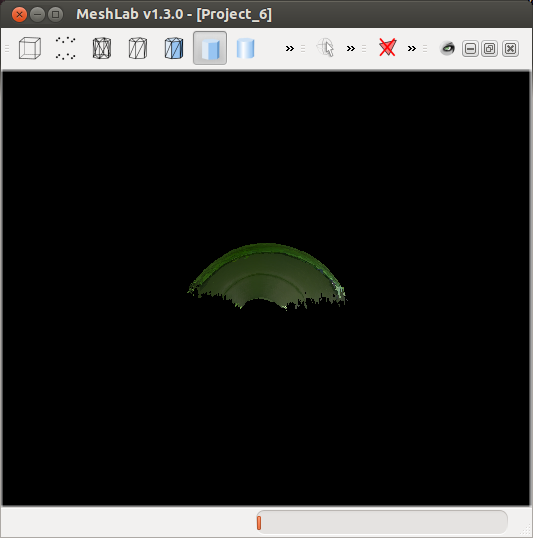}	\label{fig:teaser4}	}		\hspace*{\ImgSquizTeaserH}
		\subfloat[subfigure teaser 5][]{	\includegraphics[trim=30mm 40mm 30mm 50mm, clip=true, width=\ImgSizeMultTeaserr \textwidth]{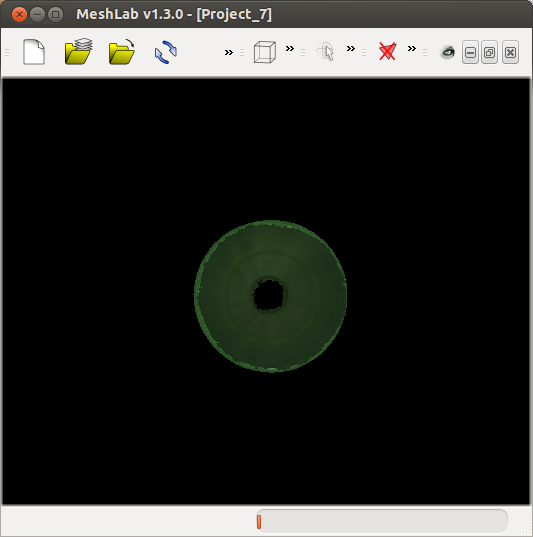}	\label{fig:teaser5}	}		\hspace*{\ImgSquizTeaserH}
		\subfloat[subfigure teaser 6][]{	\includegraphics[trim=30mm 40mm 30mm 50mm, clip=true, width=\ImgSizeMultTeaserr \textwidth]{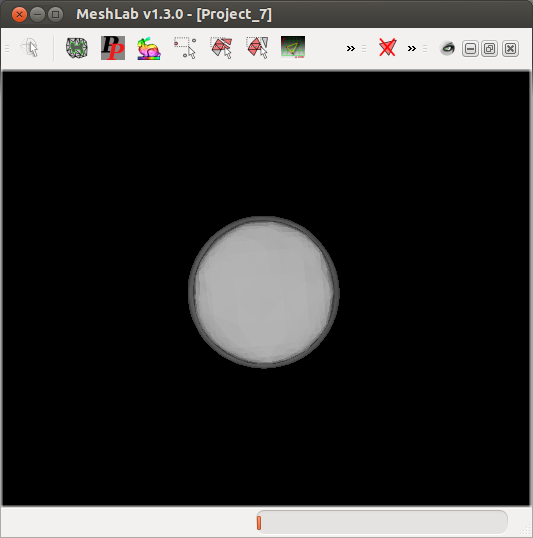}	\label{fig:teaser6}	}	
		\vspace*{-1mm}
		\caption{	Reconstruction of a symmetric, textureless object. 
				Both the front and the bottom view are provided for better visualization. 
				Left: 			Existing in-hand scanning approaches fail for such objects. 
				Middle and right: 	Successful reconstruction by the proposed in-hand scanning system that incorporates 3d hand motion capture. 
		}
		\vspace*{+4.6mm}
	\label{fig:teaser_1}
	\end{figure}

	The field has matured \cite{surveyRecentImageRegistration} since its beginning in the early 80s \cite{kanade_1981iterative} 
	and during the 90s \cite{ICP,fastProjectiveAssociation,PulliMultiview,TSDF}. Nowadays, several commercial solutions for 3D scanning with an off-the-shelf RGB-D camera have appeared, \eg, 
	Fablitec~\cite{CopyMe3D}, 
	Skanect\footnote{\url{http://skanect.occipital.com/}}, 
	iSense\footnote{\url{http://cubify.com/products/isense}}, 
	KScan3d\footnote{\url{http://www.kscan3d.com/}}, 
	Shapify~\cite{ShapifyHaoLi} and Kinect-Fusion~\cite{kinectFusionISMAR}. 
	Several open-source projects like KinFu 
	address the same problem, while 
	other commercial solutions as MakerBot-Digitizer employ a laser scanning device along with sensorimotor information from a turntable. 
	
	Instead of a turntable, an object can also be rotated by hand in case of a static camera. This setting is very convenient for hand-sized objects since moving an object is more practical than moving a camera with a cable. 
	Such a setup is also called \emph{in-hand scanning} \cite{RusinkiewiczRealTimeINHAND}. Weise \etal 
	presented a real-time in-hand scanning system \cite{weise_inHand_CVPR08} that was later augmented with online loop closure \cite{weise_inHand_CVIU11}.
	Although the results are very convincing, the method uses the hand only as a replacement of a turntable and discards the hand information. When the objects are textureless and contain very few geometric features, the in-hand scanning fails as illustrated in Figure~\ref{fig:teaser_1}.   

	In this work, we propose to use the hand motion for in-hand scanning as an additional cue to reconstruct also texturless objects. 
	Instead of discarding the hands 
	with the use of 
	a black glove~\cite{weise_inHand_CVIU11}, 
	we track the hand pose and use the captured hand motion together with texture and geometric features for object reconstruction as in Figure \ref{fig:contactCorr1}. Since the hand motion provides additional information about the object motion, we can reconstruct even textureless and symmetric objects as shown in Figure~\ref{fig:teaser_1}.


\section{Related work}\label{sec:RelatedWork}

	During the last decades several real-time in-hand scanning systems like~\cite{RusinkiewiczRealTimeINHAND,weise_inHand_CVPR08,weise_inHand_CVIU11} have been presented. 
	Such systems are able to provide a real-time registration of the input frames, while the interactivity enables the user to guide the reconstruction process.
	Assuming high temporal continuity and objects with rich geometric features, the quality of the final reconstruction can be sufficient. 
	Some methods add an offline optimization step \cite{PulliMultiview} to solve the loop closure problem, but in this case the final result might differ from the intermediate result. 
	In order to solve this issue, Weise \etal 
	presented a real-time in-hand scanning system \cite{weise_inHand_CVPR08} that was later augmented with online loop closure \cite{weise_inHand_CVIU11}.
	They follow an as-rigid-as-it-gets approach based on surfels in order to minimize registration artifacts. Due to online loop closure, the approach does not require any post-processing. 
	A different approach is proposed in STAR3D~\cite{STAR3D}. In this work a 3d level-set function is used to perform simultaneous tracking and reconstruction of rigid objects. Similarly to in-hand scanning, this approach works only for objects with sufficient geometric or texture features.   
	In order to reconstruct textureless and symmetric objects, 
	additional information from sensors, markers~\cite{ARmarkerReconstruction}, or a robotic manipulator 
	is required~\cite{kragicBirthOfTheObject,dieterFox_HandObject_Robotic}. 	
	
	In this work, we propose to extract the additional information directly from the hand within an in-hand scanning framework. 
	Instead of simply discarding the hand~\cite{RusinkiewiczRealTimeINHAND,weise_inHand_CVPR08,weise_inHand_CVIU11}, we capture the hand motion. 
	In recent years, there has been a progress in hand motion capture. In particular, capturing of hand-object interactions has become of increasing interest~\cite{JavierHandsInAction,Hamer_Hand_Manipulating,Hamer_ObjectPrior,Oikonomidis_1hand_object,kyriazis2013,LucaHands}. 
	These approaches assume that a model of the object is given, while we aim to reconstruct the object during hand-object interactions. 
	In \cite{argyros_Shape} a rigid tool is tracked in a multicamera setup to reconstruct textureless and even transparent objects. 
	Shape carving is in this case explicitly performed by the tool and the tool needs to be swept over the entire objects, which can be time-consuming. 
	In contrast to in-hand scanning, this approach needs an additional tool.    
	Static objects have also been used in \cite{slamPP} to augment a SLAM system with the pose of repetitive objects in a scene.


\section{Hand motion capture for in-hand scanning}\label{sec:AugmentedInHandScanning}

\begin{figure}[t]
		\vspace*{-2mm}
		\captionsetup[subfigure]{labelformat=empty}
		\centering
			\subfloat[subfigure 1 tracker][]{	\includegraphics[width=0.15 \textwidth]{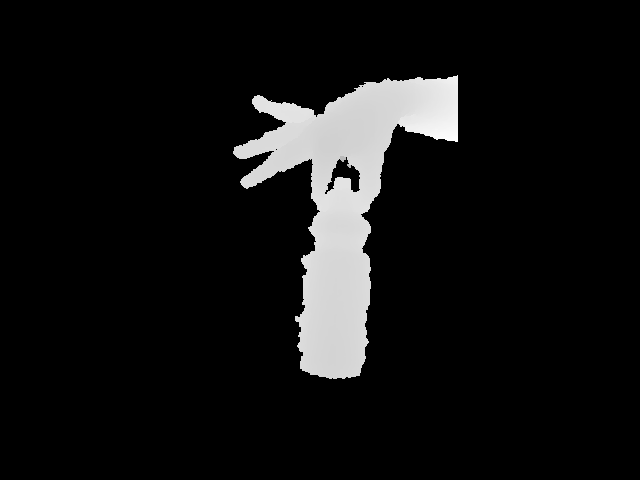}	\label{fig:tracker826_depth}	}		\hspace*{-3mm}
			\subfloat[subfigure 2 tracker][]{	\includegraphics[width=0.15 \textwidth]{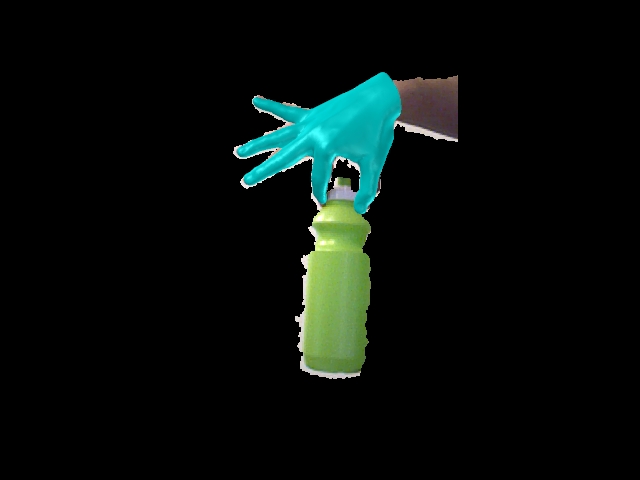}	\label{fig:tracker826_rgbdPose}	}	\hspace*{-3mm}
			\subfloat[subfigure 3 tracker][]{	\includegraphics[width=0.15 \textwidth]{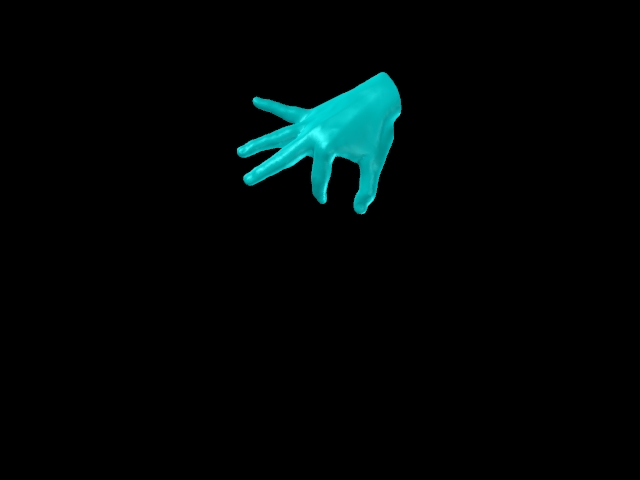}	\label{fig:tracker826_pose}	}	\\	\vspace*{-9mm}
			\subfloat[subfigure 4 tracker][]{	\includegraphics[width=0.15 \textwidth]{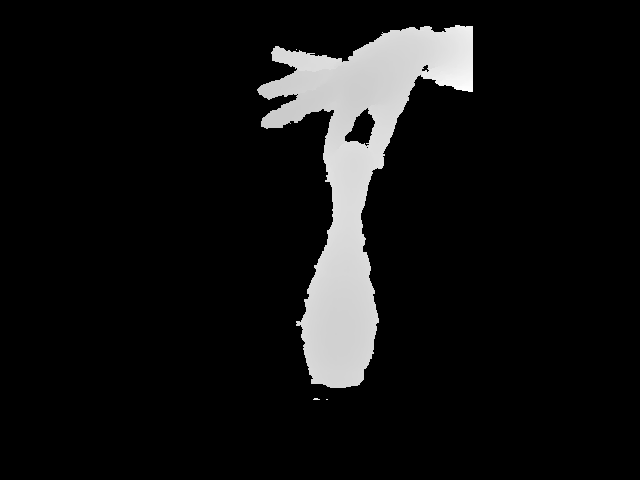}	\label{fig:tracker831_depth}	}		\hspace*{-3mm}
			\subfloat[subfigure 5 tracker][]{	\includegraphics[width=0.15 \textwidth]{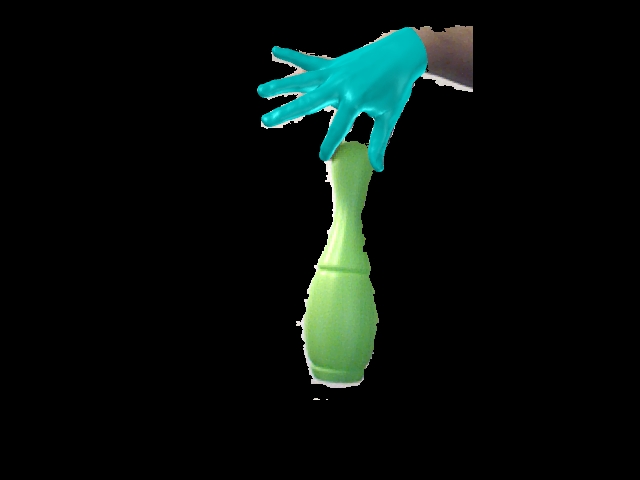}	\label{fig:tracker831_rgbdPose}	}	\hspace*{-3mm}
			\subfloat[subfigure 6 tracker][]{	\includegraphics[width=0.15 \textwidth]{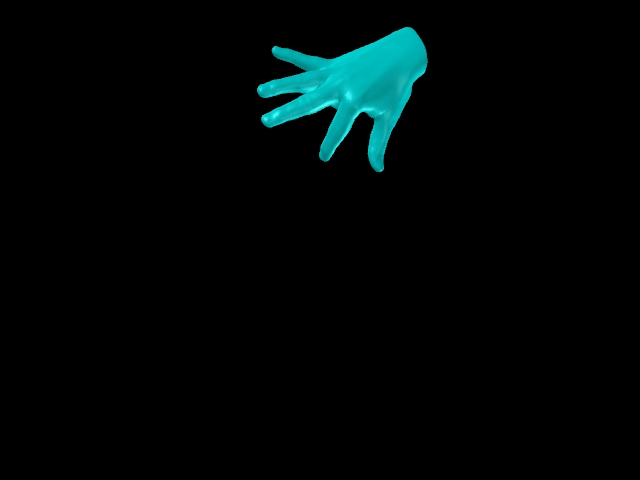}	\label{fig:tracker831_pose}	}
			\vspace*{-2mm}
		\caption{	The hand tracker used in the in-hand scanning pipeline. 
				The left image shows the raw depth input map, 
				the middle image shows the hand pose overlaid on top of the RGB-D data, while 
				the right image shows just the hand pose. 
		}
		\label{fig:handTracker}
		\end{figure}

As illustrated in Figure \ref{fig:handTracker}, we observe an RGB-D video where a hand is interacting with an object. 
The data is first preprocessed as described in Section~\ref{sec:Preprocessing} and the hand pose is estimated in each frame as described in Section \ref{sec:HandMotionCapture}. 
We then exploit the captured hand motion to reconstruct the object as shown in Figure~\ref{fig:teaser_1}. 
The reconstruction process is described Section~\ref{sec:ReconstructionMethod}.
	
\subsection{Preprocessing}\label{sec:Preprocessing}

We first remove irrelevant parts of the RGB-D image $D$ by thresholding the depth values in order to avoid unnecessary processing like normal computation for distant points. To this end, we keep only points within a specified volume. 
For the used Primesense Carmine 1.09 sensor, only points $(x,y,z)$ within the volume $[-100mm,100mm]\times[-140mm,220mm]\times[400mm,1000mm]$ are kept.     
Subsequently we apply skin color segmentation on the RGB image using the Gaussian-Mixtures-Model (GMM) of \cite{skinnColorGMM} and get the masked RGB-D images $D_o$ for the object and $D_h$ for the hand.


\subsection{Hand motion capture}\label{sec:HandMotionCapture}

In order to capture the motion of a hand, we employ an approach similar to \cite{GCPR_2014_Tzionas_Gall}. 
The approach uses a hand template mesh and parameterizes the hand pose by a skeleton and linear blend skinning~\cite{LBS_PoseSpace}. For pose estimation, we minimize an objective function, which consists of three terms:
	\begin{linenomath}
	\begin{align}\label{eq:obj}
	\begin{split}
	E(\para, D) = \quad&E_{model \rightarrow data}(\para,D_h) + \\
		           &E_{data \rightarrow model}(\para,D_h) + 
		            \gamma_c E_{collision}(\para) 
	\end{split}
	\end{align}
	\end{linenomath}
	where $D_h$ is the current preprocessed depth image for the hand and $\para$ are the pose parameters of the hand. 
	The first two terms of Equation \eqref{eq:obj} minimize the alignment error between the input depth data and the hand pose.  
	The alignment error is measured by $E_{model \rightarrow data}$, which measures how well the model fits the observed depth data, and $E_{data \rightarrow model}$, which measures how well the depth data is explained by the model. 
	$E_{collision}$ penalizes intersections of fingers and enhances realism by ensuring physically plausible poses. 
	The parameter $\gamma_c$ is set to $10$ as in \cite{GCPR_2014_Tzionas_Gall}. 
	For simplicity we do not use the additional term $E_{salient}$ of \cite{GCPR_2014_Tzionas_Gall} for the detected salient points. 
	\reviewChange{The overall hand tracking accuracy for the hand joints is approximately $17 mm$.}


\section{Object reconstruction}\label{sec:ReconstructionMethod}

In order to use the captured hand motion for 3D reconstruction, we have to infer the contact points with the object. 
This is described in Section~\ref{sec:ContactPointsComputation}. 
The reconstruction process based on the estimated hand poses and the inferred contact points is then described in Section~\ref{sec:ReconstructionMethod_PairwiseRegistration_RoughAlignment}.

\subsection{Contact Points Computation}\label{sec:ContactPointsComputation} 

\begin{figure}[htb]
	\vspace*{+1.5mm}
		\begin{center}
			\includegraphics[trim=5mm 0mm 0mm 3mm, clip=true,width=0.30 \textwidth]{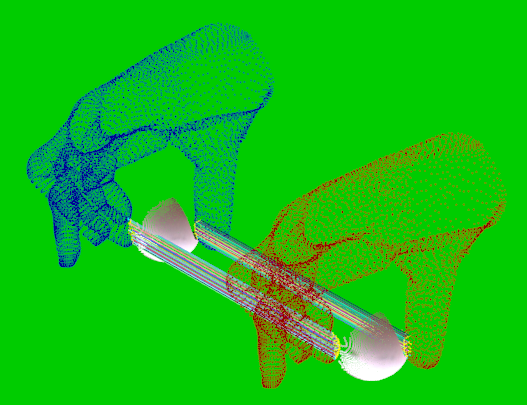}
		\end{center}
		\caption{	Illustration of the \emph{contact correspondences} $(\keyContactSourcePCL,\keyContactTargetPCL) \in \correspondencesCONTACT(\para,D_h)$ between the \emph{source frame} (red) and the \emph{target frame} (blue). 
				Although the correspondences are formed for all the vertices of the end-effectors of the manipulating fingers, we display only the detected \emph{candidate contact points} to ease visualization. 
				The candidate contact points are displayed with yellow color, while the multi-color lines show the \emph{contact correspondences}. 
				The white point cloud is a partial view of the unknown object 
				whose shape is reconstructed during hand-object interaction. 
		}
		\vspace*{-2mm}
		\label{fig:contactCorr1}
	\end{figure}

In order to compute the \emph{contact points}, we use the high-resolution mesh of the hand, which has been used for hand motion capture. 
To this end, we compute for each vertex associated to each end-effector the distance to the closest point of the object point cloud $D_o$. 
We first count for each end-effector the number of vertices with a closest distance of less than $1mm$.   
If an end-effector has more than $40$ \emph{candidate contact vertices}, it is labeled as a contact bone and all vertices of the bone are labeled as \emph{contact vertices}. 
If there are not at least 2 end-effectors selected, we iteratively increase the distance threshold by $0.5 mm$ until we have at least two end-effectors. 
In our experiments, we observed that the threshold barely exceeds $2.5mm$. 
As a result, we obtain for each frame pair the set of \emph{contact correspondences}  $(\keyContactSourcePCL,\keyContactTargetPCL) \in \correspondencesCONTACT(\para,D_h)$, where $(\keyContactSourcePCL,\keyContactTargetPCL)$ is a pair of \emph{contact vertices} in the \emph{source} and \emph{target} frame, respectively. 
Figure \ref{fig:contactCorr1} depicts the \emph{contact correspondences} for a frame pair.

\subsection{Reconstruction}\label{sec:ReconstructionMethod_PairwiseRegistration_RoughAlignment}

We use a feature-based approach for reconstruction, where we first align the currently observed point cloud \textit{(source)} to the previous frame \textit{(target)} and afterwards 
we align the transformed source by ICP to the previously accumulated transformed point cloud \cite{p2plane_metascan} for refinement.

	For pairwise registration, we combine features extracted from $D_o$ and the \emph{contact points}, which have been extracted from $D_h$ and the hand pose $\theta$. As a result, we minimize an energy function based on two weighted energies:
	\begin{linenomath}
	\begin{align}\label{eq:obj_regMatch}
	\begin{split}
		E        (\para, D_h, D_o, \RRR, \TTT) = 		\quad	&	E_{visual }(D_o, \RRR, \TTT) + \\
							      \gamma_{t}	&	E_{contact }(\para,D_h, \RRR, \TTT) 
	\end{split}
	\end{align}
	\end{linenomath}
	where $E$ is a measure of the discrepancy between the incoming and the already processed data, that needs to be minimized. 
	In that respect, we seek the rigid transformation $T=(\RRR,\TTT)$, where 
	$\RRR \in SO(3)$ 
	is a rotation matrix and 
	$\TTT \in \mathbb{R}^{3}$ 
	is a translation vector, that minimizes the energy $E$ by transforming the \emph{source frame} accordingly.

	The \emph{visual energy} $E_{visual}$ consists of two terms 
	that are computed on the visual data of the
	object point cloud
	$D_o$: 
	\begin{linenomath}
	\begin{align}\label{eq:obj_reg_visual}
	\begin{split}
	E_{visual}(D_o, \RRR, \TTT) = \quad	&E_{feat2d}(D_o, \RRR, \TTT) + \\
						&E_{feat3d}(D_o, \RRR, \TTT)
	\end{split}
	\end{align}
	\end{linenomath}
	The term $E_{feat2d}$ is based on a sparse set of correspondences $\correspondencesSIFT(D_o)$ using $2d$ SIFT \cite{sift} features that are back-projected in $3d$ by the function $\varphi(x)$: $\mathbb{R}^2\rightarrow\mathbb{R}^3$, given the intrinsic parameters of the camera.
	The $2d$ SIFT keypoint correspondences in the source and target image respectively are denoted as $(\keySiftSourceIMG,\keySiftTargetIMG) \in \correspondencesSIFT(D_o)$, 
	while $\keySiftSourcePCL=\varphi(\keySiftSourceIMG)$ and $\keySiftTargetPCL=\varphi(\keySiftTargetIMG)$ are the corresponding back-projected $3d$ points. 
	$E_{feat2d}$ is then formulated as 
	\begin{linenomath}
	\begin{align}\label{eq:obj_reg_visual_SIFT}
	\begin{split}
	E_{feat2d}&(D_o, \RRR, \TTT) =\\ &\sum_{(\keySiftSourcePCL,\keySiftTargetPCL) \in \correspondencesSIFT} \Vert \keySiftTargetPCL - (\RRR \keySiftSourcePCL + \TTT) \Vert^2.	
	\end{split}
	\end{align}
	\end{linenomath}
	
	In a similar manner, the term $E_{feat3d}$ is based on a sparse set of correspondences $\correspondencesISS(D_o)$. 
	Instead of the image domain, we operate on the $3d$ point cloud by 
	choosing ISS3D~\cite{ISS3D} keypoints and 
	the CSHOT~\cite{CSHOT} feature descriptor, that augments the 
	SHOT~\cite{SHOT} descriptor with texture information. This combination has been shown to work well for point clouds~\cite{KeypointsComparativeEvaluation_Brasil,KeypointsComparativeEvaluation_Tombari_IJCV}. 
	$E_{feat3d}$ is then formulated as 
	\begin{linenomath}
	\begin{align}\label{eq:obj_reg_visual_ISS}
	\begin{split}
	E_{feat3d}&(D_o, \RRR, \TTT) =\\ &\sum_{(\keyIssSourcePCL,\keyIssTargetPCL) \in \correspondencesISS}		\Vert \keyIssTargetPCL - (\RRR \keyIssSourcePCL + \TTT)\Vert^2.
	\end{split}
	\end{align}
	\end{linenomath}
	
	Finally, the term $E_{contact }$ depends on the current hand pose estimate $\para$ and the hand point cloud $D_h$. 
	Based on these, the \emph{contact correspondences} $\correspondencesCONTACT(\para,D_h)$ are computed as described in Section \ref{sec:ContactPointsComputation}. 
	Let $(\keyContactSourcePCL,\keyContactTargetPCL) \in \correspondencesCONTACT(\para,D_h)$ be the corresponding \emph{contact points}, \ie vertices, in the \emph{source} and \emph{target} frame respectively, 
	then $E_{contact}(\para,D_h)$ is written as 
	\begin{linenomath}
	\begin{align}\label{eq:obj_reg_contact}
	\begin{split}
	E_{contact}&(\para,D_h, \RRR, \TTT) =\\ &\sum_{(\keyContactSourcePCL,\keyContactTargetPCL) \in \correspondencesCONTACT}	\Vert \keyContactTargetPCL - (\RRR \keyContactSourcePCL + \TTT) \Vert^2.
	\end{split}
	\end{align}
	\end{linenomath}

	The two terms in the energy function \eqref{eq:obj_regMatch} 
	are weighted since 
	they
	have different characteristics. Although \emph{visual correspondences} preserve local geometric or textural details better, they tend to cause a slipping of 
	one frame upon another in case of textureless and symmetric objects. In this case, the \emph{contact correspondences} ensure that the movement of the hand is taken into account.   
	An evaluation of the weight $\gamma_{t}$ is presented in Section~\ref{sec:Experiments}.

	The sparse correspondence sets $\correspondencesSIFT$, $\correspondencesISS$, and $\correspondencesCONTACT$ provide usually an imperfect alignment of the \textit{source} frame to the \textit{target} frame   
	either because of noise, ambiguities in the visual features or the pose, or a partial violation of basic assumptions like the rigid grasping of an object during interaction. 
	For this reason, 
	we refine the aligned source frame by finding a locally optimal solution based on dense ICP~\cite{ICP} correspondences. 
	While for the sparse correspondences we align the current frame only to the previous one,
	during this refinement stage we align the current frame to the accumulation of all previously aligned frames~\cite{p2plane_metascan}, \ie the current partial reconstructed model. 
	After finding a dense set $(\keyIcpSourcePCL,\keyIcpTargetPCL) \in \correspondencesICP(D_o)$ of ICP correspondences \reviewChange{with maximum distance of $5$mm}, we  minimize the discrepancy between them 
	\begin{linenomath}
	\begin{align}\label{eq:obj_reg_icp}
	\begin{split}
	E_{icp}&(D_o, \RRR, \TTT) =\\ &\sum_{(\keyIcpSourcePCL,\keyIcpTargetPCL) \in \correspondencesICP}	\Vert \keyIcpTargetPCL - (\RRR \keyIcpSourcePCL + \TTT) \Vert^2. 
	\end{split}
	\end{align}
	\end{linenomath}

\subsubsection{Surface model}\label{sec:SurfaceModel}

	To obtain a mesh representation of the reconstructed object,  
	we first employ a truncated signed distance function (TSDF)~\cite{TSDF,kinectFusionISMAR} to get a volumetric representation. 
	The TSDF volume has a dimension of $350 mm$ for all objects with $256$ voxel resolution and $6 mm$ maximum voxel size. 
	Subsequently we apply the marching-cubes~\cite{MarchingCubes} method to extract a mesh and remove tiny disconnected components.
	The final mesh is then obtained by Laplacian smoothing~\cite{Meshlab_HC_LaplacianSmoothing} followed by Poisson reconstruction~\cite{Meshlab_PoissonReconstruction} with an octree 
	with $10$ layers in order to get a smooth, water-tight mesh with preserved details.


\section{Experiments}\label{sec:Experiments}

	In this section we show that although existing in-hand scanning pipelines fail for symmetric and textureless objects,  
	the incorporation of hand motion capture can effectively improve the reconstruction, enabling the efficient and full reconstruction of such objects without the use of additional intrusive markers or devices in the scene. 
	We present thus for the first time the effective reconstruction of $4$ symmetric objects with an in-hand scanning system, 
	which cannot be reconstructed by two state-of-the-art reconstruction systems. 
	Furthermore, we perform an experiment with synthetic data, showing that the pipeline can also be applied to multicamera RGB videos.  
	
	The recorded sequences, calibration data, hand motion data, as well as 
	video results, the resulting meshes and the source code for reconstruction 
	are publicly available\footnote{\url{http://files.is.tue.mpg.de/dtzionas/ihScanning}}.

	\newcommand{\ImgSquizSynthH}{-05mm}
	\newcommand{\ImgSquizSynthV}{-03mm}
	\newcommand{\ImgSizeMultSynth}{0.24}	
	\begin{figure}[t]
	\vspace*{-2mm}
	\captionsetup[subfigure]{}
	\centering
		\subfloat[subfigure synthetic small hands NOO][]{	\includegraphics[trim=20mm 30mm 30mm 25mm, clip=true, width=\ImgSizeMultSynth \textwidth]{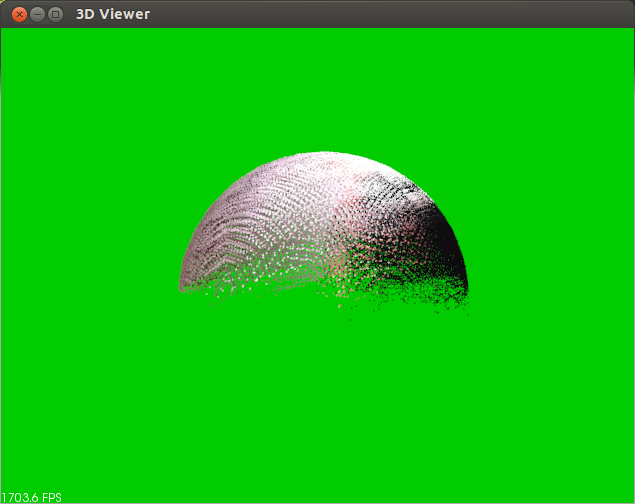}	\label{fig:synthetic_small_handsNOO}	}		\hspace*{\ImgSquizSynthH}
		\subfloat[subfigure synthetic small hands YES][]{	\includegraphics[trim=20mm 30mm 30mm 25mm, clip=true, width=\ImgSizeMultSynth \textwidth]{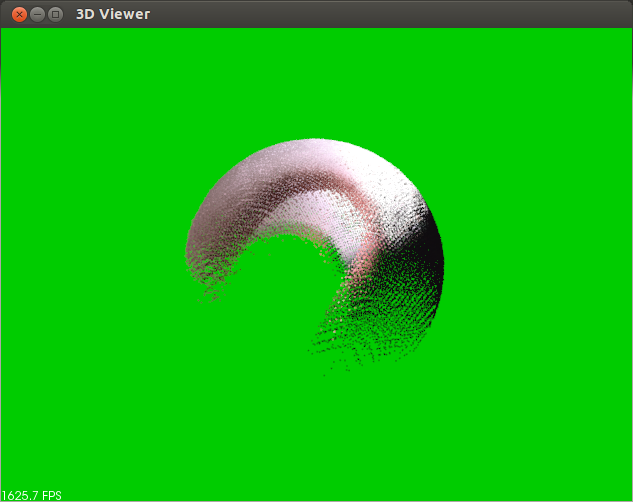}	\label{fig:synthetic_small_handsYES}	}	
		\vspace*{1mm}
		\caption{
				\reviewChange
				{
				Reconstruction without (a) and with (b) hand motion capture on synthetic data generated from \cite{LucaHands}. 
				Since the \emph{visual correspondences} alone are not descriptive enough for symmetric objects, the reconstruction collapses to a hemisphere (a). 
				On the contrary, the use of hand motion capture gives meaningful \emph{contact correspondences}, successfully driving the reconstruction process (b). 
				The clear observation of the occlusions by the manipulating fingers (b) indicates a sensible registration. 
				The motion includes some notable translation and rotation, but the object is not fully rotated in order to allow for a complete reconstruction. 
				}
		}
	\label{fig:synthetic1}
	\end{figure}

\begin{table*}[ht]
		\scriptsize 
		\begin{center}
			\caption{\reviewChange{Quantitative evaluation of the captured object shapes. The ground-truth parameters, estimated parameters and errors are given. We compare our proposed setup with $\gamma_t=15$ with the methods KinFu and Skanect. 
			For the methods highlighted with (*), we perform a reconstruction three times and we report the best results of them. 
					}
			}
			\vspace*{1mm}
			\label{fig:QResults}
			\setlength{\tabcolsep}{1pt}	
			\begin{tabular}{|l|c|c|c|c|c|c|c|c|c|c|c|c|c|c|c|c|c|c|c|}																																																																									\hhline{-~-~--~--~--~--~--~~}
			
				\multicolumn{1}{|c|}{~~Dimensions Comparison~~} 		& 
				\multicolumn{1}{ c|}{} 						& 
				~G.Truth~							& 	&
				\multicolumn{2}{c|}{Ours $\gamma_t=15$}			& 	&
				\multicolumn{2}{c|}{\reviewChange{KinFu (*)}}			& 	&
				\multicolumn{2}{c|}{\reviewChange{Skanect (*)}}		& 	&
				\multicolumn{2}{c|}{\reviewChange{Detect.Baseline}}		& 	&
				\multicolumn{2}{c|}{\reviewChange{Enriched Texture}}		& 	 
				\multicolumn{1}{c}{}						& 
				\multicolumn{1}{c}{}																																																																												\\	\hhline{-~-~--~--~--~--~--~~}
				\multicolumn{1}{c}{} 						& 
				\multicolumn{1}{c}{} 						& 
				\multicolumn{1}{c}{}						& 	&
				~Capture~							& 
				~Diff.~								& 	&
				~Capture~							& 
				~Diff.~								& 	&
				~Capture~							& 
				~Diff.~								& 	&
				~Capture~							& 
				~Diff.~								& 	&
				~Capture~							& 
				~Diff.~								& 	
				\multicolumn{1}{c}{}						& 
				\multicolumn{1}{c}{} 																																																																												\\	\hhline{~~~~--~--~--~--~--~~}
				\noalign{\smallskip}																																																																													\hhline{-~-~--~--~--~--~~~~-}
				
							{~Water-bottle 		diameter~}	& {}			& 	{~$73$ }		& {}			& { $82.3$ }		& {~$9.3$~}				& {}			& { $66.2$ }		& {~$6.8$~}				& {}			& { $64.3$ }		& {~$8.7$~}				& {}			& { $86.6$ }		& {~$13.6$~}				& \multicolumn{1}{c}{}	& \multicolumn{1}{c}{}	& \multicolumn{1}{c}{}		& {} 	& \multirow{9}{*}{\centering\begin{turn}{90}mm\end{turn}}	\\	\hhline{-~-~--~--~--~--~~~~~}
							{~Water-bottle 		height~}	& {}			& 	{~$218$~}		& {}			& {~$225.4$~}		& {~$7.4$~}				& {}			& { $195.7$ }		& {~$22.3$~}				& {}			& { $222.1$ }		& {~$4.1$~}				& {}			& { $237.4$ }		& {~$19.4$~}				& \multicolumn{1}{c}{}	& \multicolumn{1}{c}{}	& \multicolumn{1}{c}{}		& {} 	& {} 								\\	\hhline{-~-~--~--~--~--~--~~}
							{~Bowling-pin head 	diameter~}	& {}			& 	{~$50$ }		& {}			& { $50.8$ }		& { $0.8$ }				& {}			& { $54.1$ }		& {~$4.1$~}				& {}			& { $39.0$ }		& {~$11.0$~}				& {}			& { $48.7$ }		& {~$1.3$~}				& {}			& { $49.8$ }		& {~$0.2$~}			& {} 	& {} 								\\	\hhline{-~-~--~--~--~--~--~~}
							{~Bowling-pin body   	diameter~}	& {}			& 	{ $82$ }		& {}			& { $90.0$ }		& {~$8.0$~}				& {}			& { $70.9$ }		& {~$11.1$~}				& {}			& { $63.8$ }		& {~$18.2$~}				& {}			& { $93.2$ }		& {~$11.2$~}				& {}			& { $89.4$ }		& {~$7.4$~}			& {} 	& {} 								\\	\hhline{-~-~--~--~--~--~--~~}
							{~Bowling-pin  		height~}	& {}			& 	{~$268$~}		& {}			& {~$275.2$~}		& {~$7.2$~}				& {}			& { $239.3$ }		& {~$28.7$~}				& {}			& { $270.9$ }		& {~$2.9$~}				& {}			& { $272.4$ }		& {~$4.4$~}				& {}			& { $267.7$ }		& {~$0.3$~}			& {}	& {} 								\\	\hhline{-~-~--~--~--~--~--~~}
							{~Small-bottle 		diameter~}	& {}			& 	{~$52$ }		& {}			& { $57.7$ }		& { $5.7$ }				& {}			& { $45.6$ }		& {~$6.4$~}				& {}			& { $39.5$ }		& {~$12.5$~}				& {}			& { $61.6$ }		& {~$9.6$~}				& \multicolumn{1}{c}{}	& \multicolumn{1}{c}{}	& \multicolumn{1}{c}{}		& {} 	& {}  								\\	\hhline{-~-~--~--~--~--~~~~~}
							{~Small-bottle 		height~}	& {}			& 	{~$80$ }		& {}			& { $89.5$ }		& { $9.5$ }				& {}			& { $78.1$ }		& {~$1.9$~}				& {}			& { $84.9$ }		& {~$4.9$~}				& {}			& { $95.0$ }		& {~$15.0$~}				& \multicolumn{1}{c}{}	& \multicolumn{1}{c}{}	& \multicolumn{1}{c}{}		& {} 	& {}  								\\	\hhline{-~-~--~--~--~--~~~~~}
							{~Sphere 		diameter~}	& {}			& 	{~$70$ }		& {}			& { $71.4$ }		& { $1.4$ } 				& {}			& { $46.9$ }		& {~$23.1$~}				& {}			& { $43.8$ }		& {~$26.2$~}				& {}			& { $72.2$ }		& {~$2.2$~}				& \multicolumn{1}{c}{}	& \multicolumn{1}{c}{}	& \multicolumn{1}{c}{}		& {} 	& {}  								\\	\hhline{-~-~--~--~--~--~~~~~}
							{~Average~}				& \multicolumn{1}{c}{}	& 	\multicolumn{1}{c}{}	& \multicolumn{1}{c}{}	& \multicolumn{1}{c}{}	& \multicolumn{1}{|c|}{ $6.1625$ }	& \multicolumn{1}{c}{}	& \multicolumn{1}{c}{}	& \multicolumn{1}{|c|}{ $13.05$ }	& \multicolumn{1}{c}{}	& \multicolumn{1}{c}{}	& \multicolumn{1}{|c|}{ $11.0625$ }	& \multicolumn{1}{c}{}	& \multicolumn{1}{c}{}	& \multicolumn{1}{|c|}{$9.5875$} 	& \multicolumn{1}{c}{}	& \multicolumn{1}{c}{}	& \multicolumn{1}{c}{} 		& {} 	& {}								\\	\hhline{-~~~~-~~-~~-~~-~~~~-}
				\noalign{\smallskip}																																																																													\hhline{-~-~--~--~--~--~~~~-}
							{~Sphere 		volume~}	& {}			& 	{~$179503$ }		& {}			& { $190490$ }		& { $10987$ }				& {}			& { $53988$ }		& { $125515$ } 				& {}			& { $43974$ }		& { $135529$ } 				& {}			& { $196965$ }		& {~$17462$~}				& \multicolumn{1}{c}{}	& \multicolumn{1}{c}{}	& \multicolumn{1}{c}{}		& {} & {$mm^3$}  							\\	\hhline{-~-~--~--~--~--~~~~-}
			\end{tabular}
		\end{center}
		\vspace*{-1mm}
\end{table*}
	
\subsection{Synthetic data}\label{sec:Experiments_SyntheticData}

	In order to generate synthetic data we use the publicly available\footnote{\url{http://cvg.ethz.ch/research/ih-mocap}} data of a multicamera RGB hand tracker \cite{LucaHands}. We use the frames $180$-$203$ 
	of the sequence in which a hand interacts with a rigid ball. 
	We generate synthetic point clouds by rendering the moving meshes. 
	We then apply the pipeline described in Section \ref{sec:ReconstructionMethod} to the rendered point clouds and use the hand meshes and motion data of \cite{LucaHands}. 
	
	The resulting accumulated and aligned point cloud is depicted in Figure~\ref{fig:synthetic1}. 
	Figure \ref{fig:synthetic1}(a) 
	shows the reconstruction without hand motion data, while 
	Figure \ref{fig:synthetic1}(b) 
	shows the reconstruction after the incorporation of hand motion into the in-hand scanning system. 
	The reconstruction without the hand motion data collapses to a degenerate hemisphere, while 
	it is clear that 
	the hand motion data significantly contributes towards the effective reconstruction of the manipulated object. Parts of the object are never visible in the sequence.  
	The occlusions caused by the manipulating fingers can be clearly observed, verifying the correct registration of the camera frames.

	\begin{figure}[t]
	\vspace*{+1mm}
		\begin{center}
			\includegraphics[trim=0mm 25mm 0mm 10mm, clip=true,width=0.45 \textwidth]{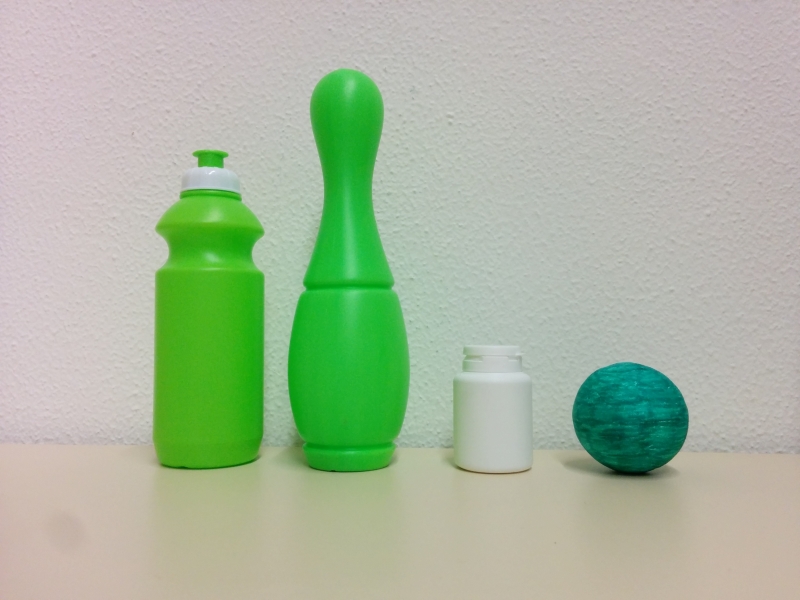}
		\end{center}
		\caption{	The objects to be reconstructed 
				(left to right):
				a \emph{water-bottle}, 
				a \emph{bowling-pin}, 
				a \emph{small-bottle} and 
				a \emph{sphere}. 
				The dimensions of the objects are summarized in Table \ref{fig:QResults}. 
				All four objects are characterized by high symmetry and lack of distinctive geometrical and textural features, causing 
				existing in-hand pipelines to fail. We perform successful reconstruction of all four objects. 
		}
		\vspace*{-2mm}
		\label{fig:photo_objectsAll}
	\end{figure}

\subsection{Realistic data}\label{sec:Experiments_RealisticData}

	For our experiments with realistic data, we use a Primesense Carmine 1.09 short-range, structured-light RGB-D camera. 
	Structured light sensors may not be optimal for hand pose estimation, in contrast to time-of-flight sensors, because significant parts of the hand completely disappear from the depth image 
	in case of reflections 
	or at some viewing angles. Nevertheless, the used hand tracker worked well with the sensor.  
	
	In order to perform both a qualitative and a quantitative evaluation, we have captured new sequences for the four objects depicted in Figure \ref{fig:photo_objectsAll}. 
	As seen in Table \ref{fig:QResults}, the size of the objects varies in order to be representative of several everyday objects. 
	However, all objects have in common the high symmetry and the lack of distinctive geometrical and textural features, that renders them especially challenging for existing in-hand scanning systems. 
	
	In the following we show the successful reconstruction of these objects for the first time with an in-hand scanning system, while we systematically evaluate the performance 
	of our pipeline both with respect to existing baselines, ground-truth object dimensions as well as state-of-the-art systems.

\subsubsection{Quantitative evaluation}\label{sec:Experiments_RealisticData_Quantitative}

	Acquiring a ground-truth measure is difficult for most objects, however, for symmetric ones it is easy to measure the dimensions of some distinctive areas. 
	We therefore measure manually the distinctive dimensions of the objects depicted in Figure \ref{fig:photo_objectsAll} in order to quantitatively evaluate the proposed setup. 
	The ground-truth dimensions, along with the measured ones by our approach and the measurement error, are presented in Table \ref{fig:QResults}. 
	Especially for the case of the sphere, we can easily acquire a ground truth value for its volume, that is less prone to measurement errors introduced by human factors. 
	
\begin{figure}[t]
	\captionsetup[subfigure]{labelformat=empty}
	\centering
		\subfloat[subfigure 1 errorGraphWeights][]{ \includegraphics[trim=00mm 00mm 00mm 10mm, clip=true, width=0.45 \textwidth]{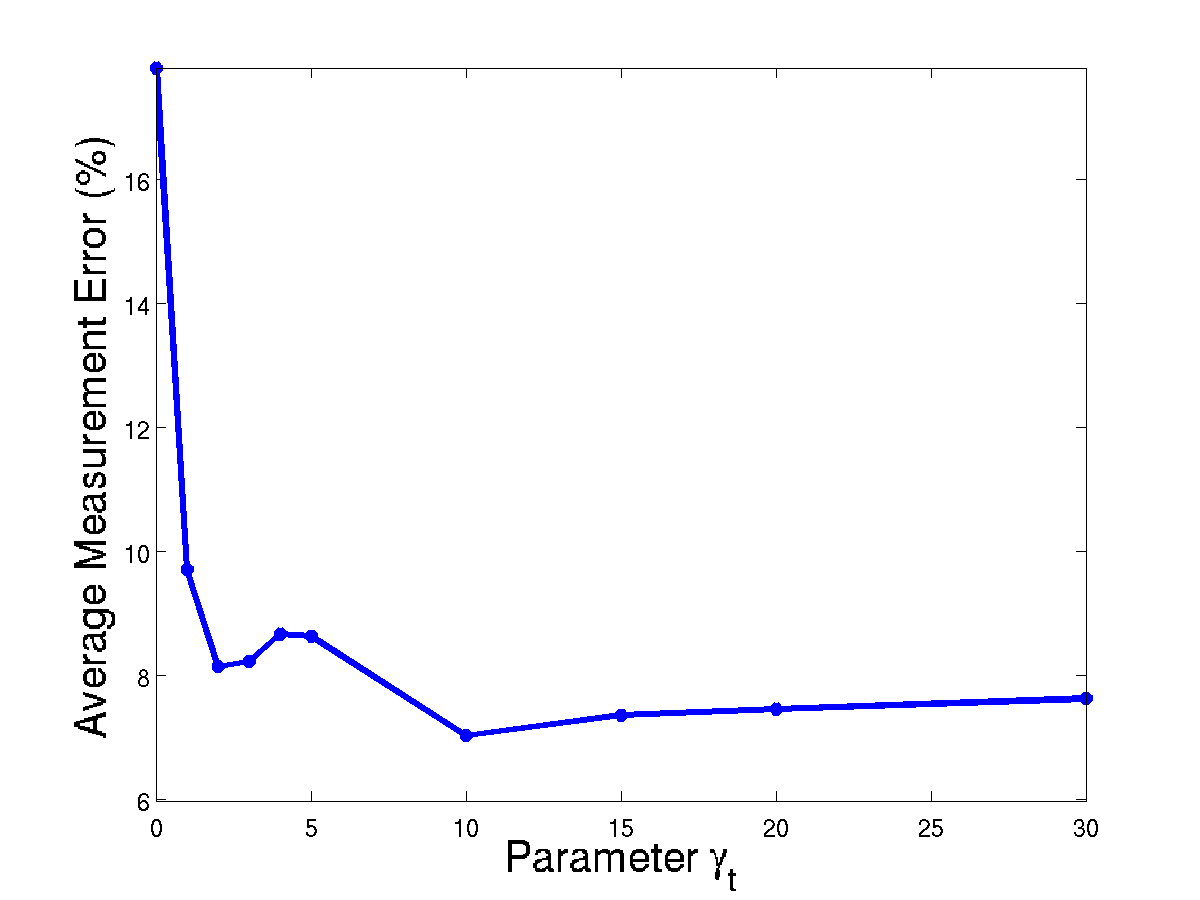} 
		}
		\vspace*{-3mm}
	\caption{	Quantitative evaluation of the weight $\gamma_t$ of the energy function \eqref{eq:obj_regMatch} based on the ground truth dimensions of the objects presented in Table \ref{fig:QResults}. The error for each parameter is normalized by the ground-truth value. 
	}
	\label{fig:weightsGammaT}
	\end{figure}

	During quantitative evaluation, we measured the distinctive dimensions of the water-tight meshes that are reconstructed by our pipeline. 
	We then evaluate the most important parameter of our pipeline, namely the weight $\gamma_t$ that steers the influence of the \emph{contact correspondences} in the in-hand scanning system. 
	The results of our experiments are summarized in Figure \ref{fig:weightsGammaT}, which plots the mean accumulated estimation errors of the eight parameters $p$ normalized by the ground-truth values, \ie, $\frac{1}{8} \sum_p \frac{\vert p_{est} - p_{GT} \vert}{p_{GT}}$. 
	Without the \emph{contact correspondences} and using only \emph{visual features} ($\gamma_t=0$), the error is relatively high since the reconstruction for symmetric objects fails in this case. 
	Even small values for $\gamma_t$ result in an abrupt drop in the error metric, however, the influence of the \emph{contact correspondences} starts becoming more apparent for values above $5$. In all subsequent experiments, we use $\gamma_t=15$.

	The performance of our setup is described in more details in Table \ref{fig:QResults} which provides a direct comparison to the measured object dimensions. 
	The average error is only $6 mm$, comparable to the noise of commodity RGB-D sensors, showing the potential of such a system for a wide range of everyday applications.

		\newcommand{\ImgSquizStickersH}{-04mm}
		\newcommand{\ImgSquizStickersV}{-08mm}
		\newcommand{\ImgSizeMultStickers}{0.16}		
		\begin{figure}[t!]
		\vspace*{-1.0mm}
		\captionsetup[subfigure]{labelformat=empty}
		\centering
			\subfloat[subfigure BowlingPin_WITH_STICKERS 1][]{	\includegraphics[trim=50mm 05mm 45mm 05mm, clip=true, height=\ImgSizeMultStickers  \textwidth]{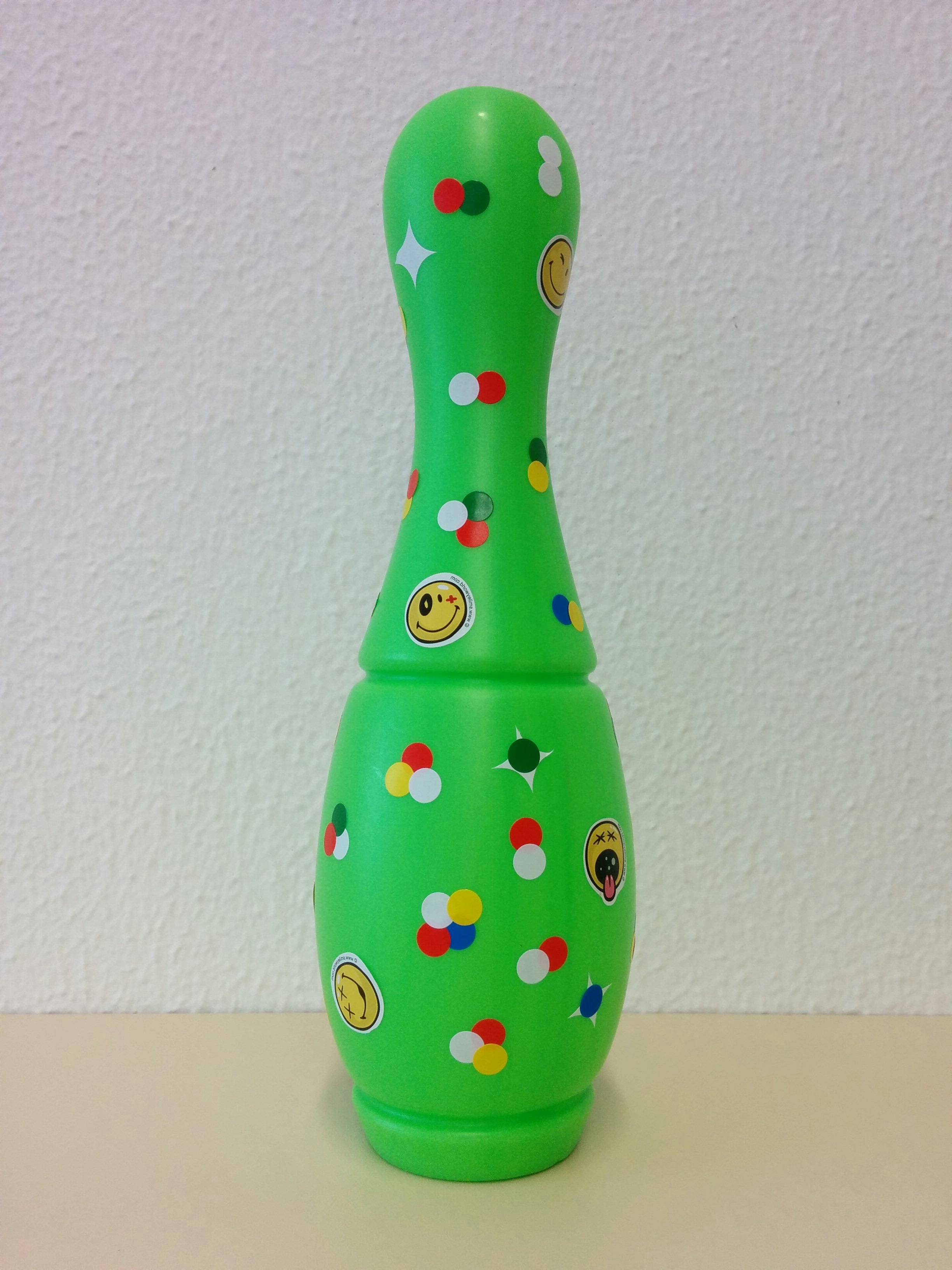}		\label{fig:BowlingPin_WITH_STICKERS_real}	}		\hspace*{\ImgSquizStickersH}
			\subfloat[subfigure BowlingPin_WITH_STICKERS 2][]{	\includegraphics[trim=45mm 13mm 45mm 30mm, clip=true, height=\ImgSizeMultStickers  \textwidth]{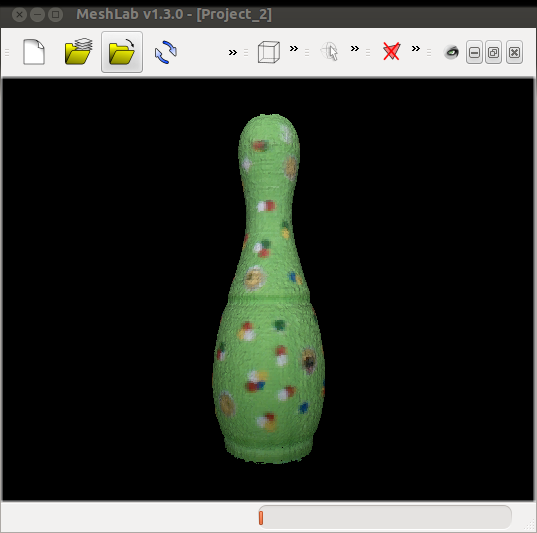}		\label{fig:BowlingPin_WITH_STICKERS_tsdf}	}		\hspace*{\ImgSquizStickersH}
			\subfloat[subfigure BowlingPin_WITH_STICKERS 3][]{	\includegraphics[trim=45mm 13mm 45mm 30mm, clip=true, height=\ImgSizeMultStickers  \textwidth]{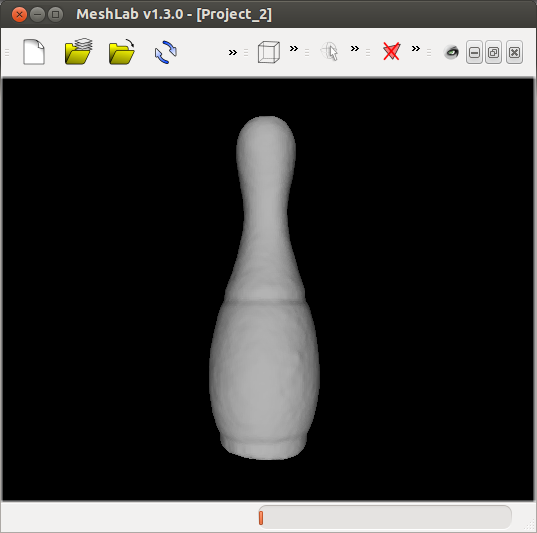}		\label{fig:BowlingPin_WITH_STICKERS_poisson}	}		
			\vspace*{-1mm}
			\caption{	
					\reviewChange
					{
					The \emph{bowling-pin} object enriched with 2d texture using stickers (left). The added texture allows for the reconstruction of a reference ground truth shape (middle, right) facilitating quantitative evaluation. 
					}
			}
		\label{fig:BowlingPin_WITH_STICKERS}
		\end{figure}
	
	\reviewChange
	{
	For evaluation against a reference reconstructed shape, we also add textural features on the \emph{bowling-pin} in the form of stickers, as depicted in Figure \ref{fig:BowlingPin_WITH_STICKERS}, without altering the geometrical shape of the object. 
	We then perform the reconstruction by rotating the object on a turntable.   
	The resulting reconstruction is illustrated in Figure \ref{fig:BowlingPin_WITH_STICKERS}, while quantitative measures are provided for comparison in Table \ref{fig:QResults}. 
	}

	We further test the effectiveness of our system in comparison to two state-of-the-art systems, namely 
	\emph{KinFu}\footnote{\url{http://pointclouds.org/documentation/tutorials/using\_kinfu\_large\_scale.php}}, an open-source implementation of Kinect-Fusion \cite{kinectFusionISMAR}, and 
	the similar commercial system 
	\emph{Skanect}\footnote{\url{http://skanect.occipital.com}}. 
	For technical reasons, we use KinFu with a Kinect and Skanect with the Structure-IO camera. 
	Existing in-hand scanning approaches are expected to have a performance similar to these systems.
	For comparison, we use \emph{KinFu} and \emph{Skanect} to reconstruct the four objects depicted in Figure \ref{fig:photo_objectsAll}. 
	A turntable rotates each object for approximately $450$ degrees in front of a static camera, while we repeat the process three times and report only the best run in order to assure objectiveness. 
	\reviewChange{Quantitative performance measures are provided for these methods in Table \ref{fig:QResults}.}

	\reviewChange
	{
	In order to show the important role of the hand pose in our reconstruction pipeline, we replace the contact correspondences $\correspondencesCONTACT$ based on \emph{contact vertices} 
	with correspondences $\correspondencesDETECTOR$ based on a \emph{contact detector}. 
	In that respect we train a Hough forest \cite{juergen_Hough} detector that detects finger-object contacts in RGB images. 
	We then establish correspondences $(X_{det},X_{det}') \in \correspondencesDETECTOR$ between the points enclosed by the detection bounding boxes in the \emph{source} and \emph{target} frames simply by associating points with the same 2d coordinates inside the fixed-sized bounding boxes. 
	In that respect, the term $E_{contact}(\para,D_h, \RRR, \TTT)$ in the objective function \eqref{eq:obj_regMatch} is replaced by the term 
	\begin{linenomath}
	\begin{align}\label{eq:obj_reg_detector}
	\begin{split}
	E_{detector}&(D_h, D_o, \RRR, \TTT) =\\ &\sum_{(\keyDetectorSourcePCL,\keyDetectorTargetPCL) \in \correspondencesDETECTOR}	\Vert \keyDetectorTargetPCL - (\RRR \keyDetectorSourcePCL + \TTT) \Vert^2.
	\end{split}
	\end{align}
	\end{linenomath}
	The results of the reconstruction, depicted in Figure \ref{fig:BaselineDetector_CorrRaster2d_results}, show that the reconstruction is either incomplete or it has major flaws, which is supported by the numbers in Table \ref{fig:QResults}.   
	}
	
	\reviewChange
	{
	In order to measure the accuracy of the contact correspondences obtained by the \emph{hand pose} or the \emph{contact detector}, 
	we manually annotated two points for each of the two manipulating fingers for pairs of consecutive frames and we do so for every 10th frame in our four sequences. We then measure the pairwise registration error for each annotated pair $(X_{gt},X'_{gt})$ by $\Vert X'_{gt}- (\RRR  X_{gt} + \TTT )\Vert$.
	The results are summarized in Table \ref{fig:QResultsDETECTOR} and show that the hand tracker is more accurate for pairwise registration than a detection based approach.  
	}
	
		\newcommand{\ImgSquizWeightsH}{-05mm}
		\newcommand{\ImgSquizWeightsV}{-08mm}
		\newcommand{\ImgSizeMultWeights}{0.12}
		\begin{figure*}[t]
			\vspace*{-3mm}
			\captionsetup[subfigure]{labelformat=empty}
			\centering
				\subfloat[subfigure Weights 826 00][]{	\includegraphics[trim=00mm 20mm 00mm 30mm, clip=true, height=\ImgSizeMultWeights \textwidth]{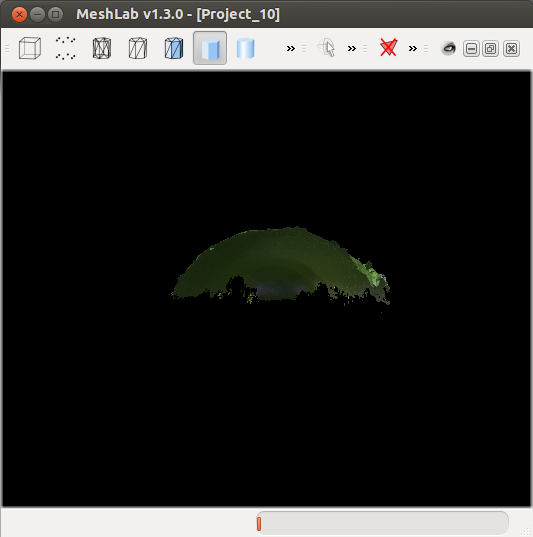}	\label{fig:Weights_826_00}	}		\hspace*{\ImgSquizWeightsH}
				\subfloat[subfigure Weights 826 01][]{	\includegraphics[trim=00mm 20mm 00mm 30mm, clip=true, height=\ImgSizeMultWeights \textwidth]{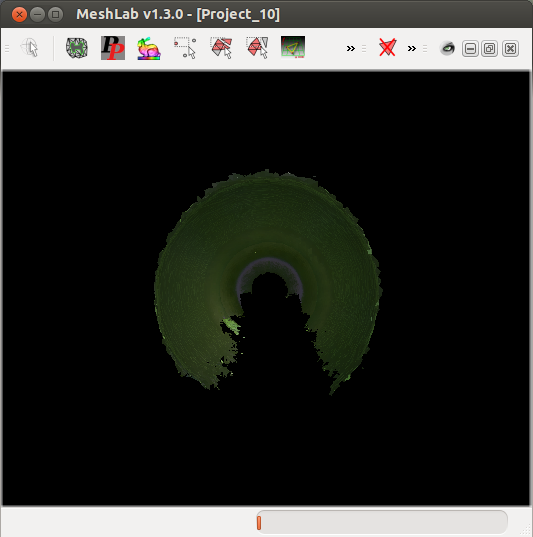}	\label{fig:Weights_826_01}	}		\hspace*{\ImgSquizWeightsH}
				\subfloat[subfigure Weights 826 05][]{	\includegraphics[trim=00mm 20mm 00mm 30mm, clip=true, height=\ImgSizeMultWeights \textwidth]{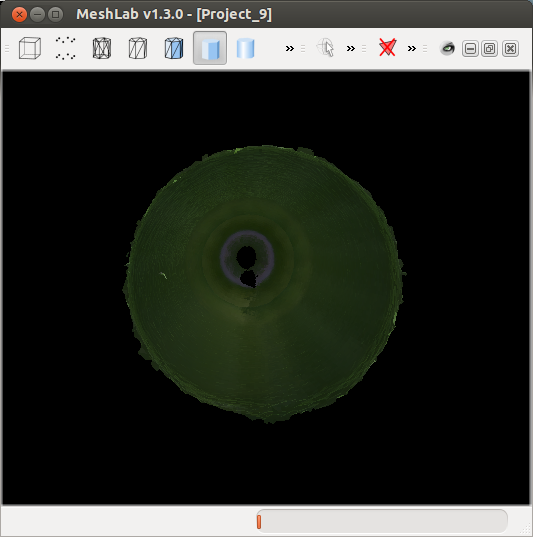}	\label{fig:Weights_826_05}	}		\hspace*{\ImgSquizWeightsH}
				\subfloat[subfigure Weights 826 10][]{	\includegraphics[trim=00mm 20mm 00mm 30mm, clip=true, height=\ImgSizeMultWeights \textwidth]{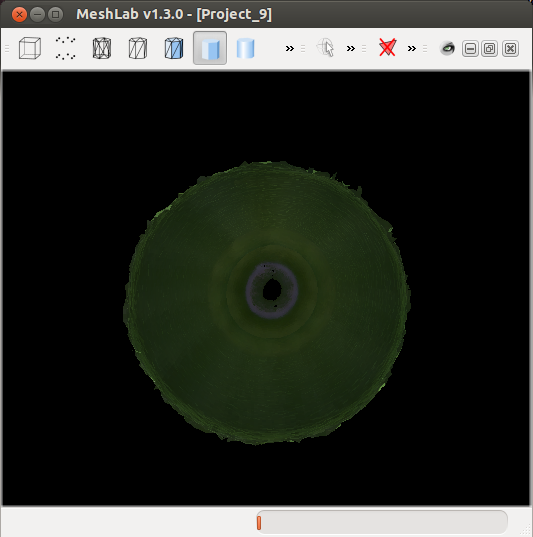}	\label{fig:Weights_826_10}	}		\hspace*{\ImgSquizWeightsH}
				\subfloat[subfigure Weights 826 15][]{	\includegraphics[trim=00mm 20mm 00mm 30mm, clip=true, height=\ImgSizeMultWeights \textwidth]{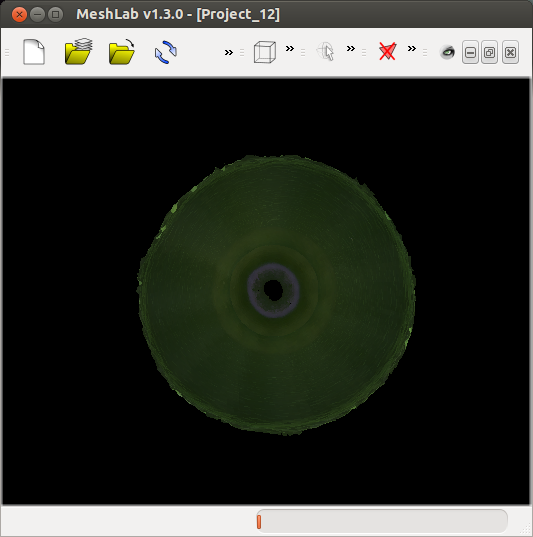}	\label{fig:Weights_826_15}	}		\hspace*{\ImgSquizWeightsH}
				\subfloat[subfigure Weights 826 20][]{	\includegraphics[trim=00mm 20mm 00mm 30mm, clip=true, height=\ImgSizeMultWeights \textwidth]{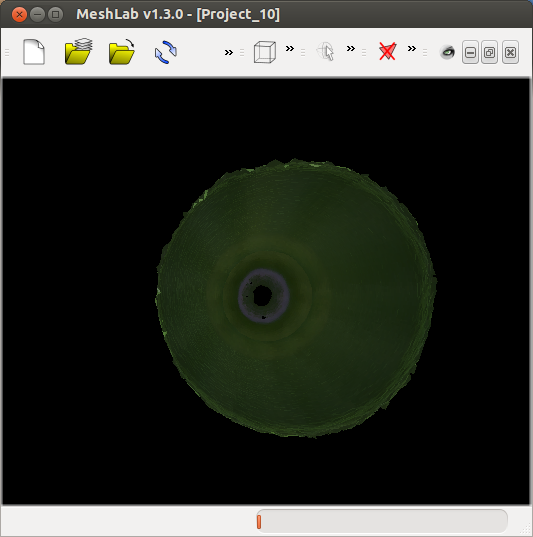}	\label{fig:Weights_826_20}	}	\\	\vspace*{\ImgSquizWeightsV}
				
				\subfloat[subfigure Weights 831 00][]{	\includegraphics[trim=00mm 20mm 00mm 30mm, clip=true, height=\ImgSizeMultWeights \textwidth]{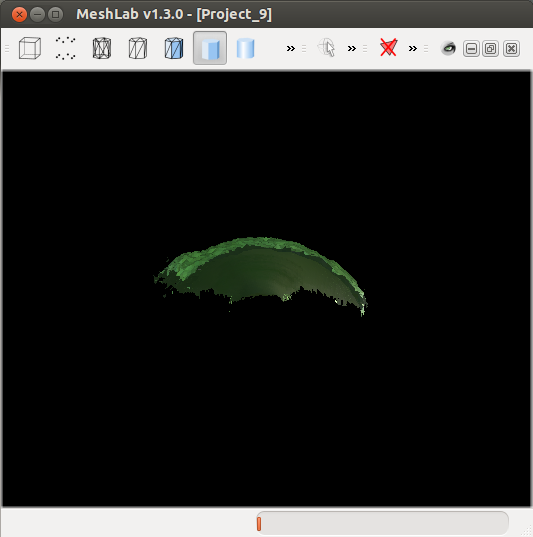}	\label{fig:Weights_831_00}	}		\hspace*{\ImgSquizWeightsH}
				\subfloat[subfigure Weights 831 01][]{	\includegraphics[trim=00mm 20mm 00mm 30mm, clip=true, height=\ImgSizeMultWeights \textwidth]{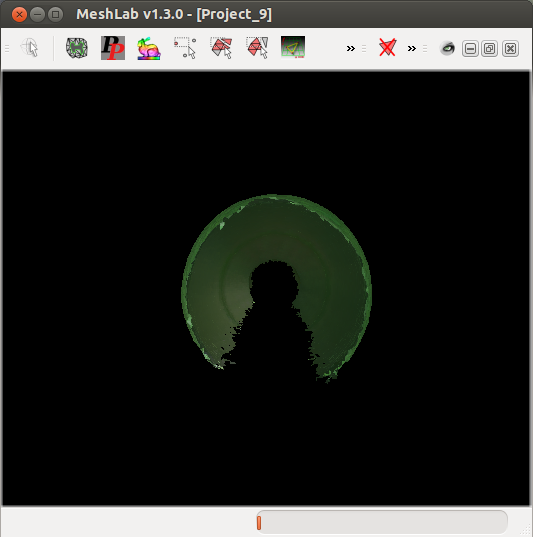}	\label{fig:Weights_831_01}	}		\hspace*{\ImgSquizWeightsH}
				\subfloat[subfigure Weights 831 05][]{	\includegraphics[trim=00mm 20mm 00mm 30mm, clip=true, height=\ImgSizeMultWeights \textwidth]{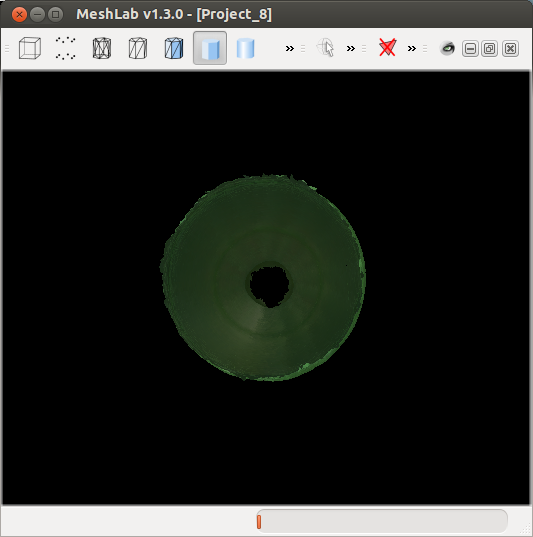}	\label{fig:Weights_831_05}	}		\hspace*{\ImgSquizWeightsH}
				\subfloat[subfigure Weights 831 10][]{	\includegraphics[trim=00mm 20mm 00mm 30mm, clip=true, height=\ImgSizeMultWeights \textwidth]{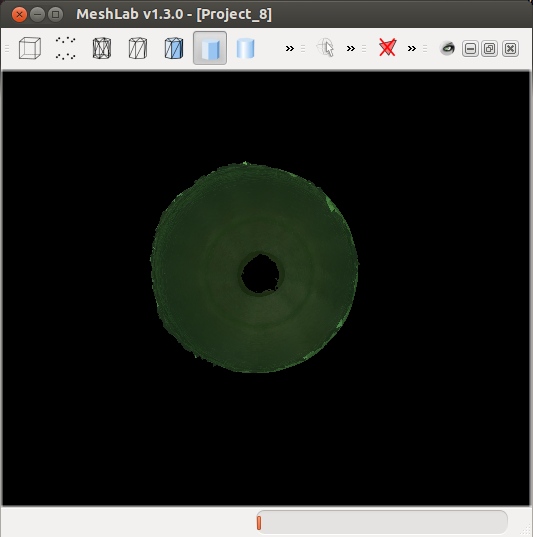}	\label{fig:Weights_831_10}	}		\hspace*{\ImgSquizWeightsH}
				\subfloat[subfigure Weights 831 15][]{	\includegraphics[trim=00mm 20mm 00mm 30mm, clip=true, height=\ImgSizeMultWeights \textwidth]{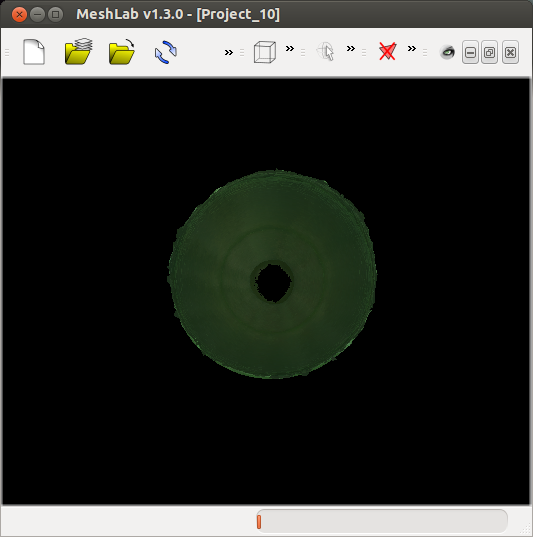}	\label{fig:Weights_831_15}	}		\hspace*{\ImgSquizWeightsH}
				\subfloat[subfigure Weights 831 20][]{	\includegraphics[trim=00mm 20mm 00mm 30mm, clip=true, height=\ImgSizeMultWeights \textwidth]{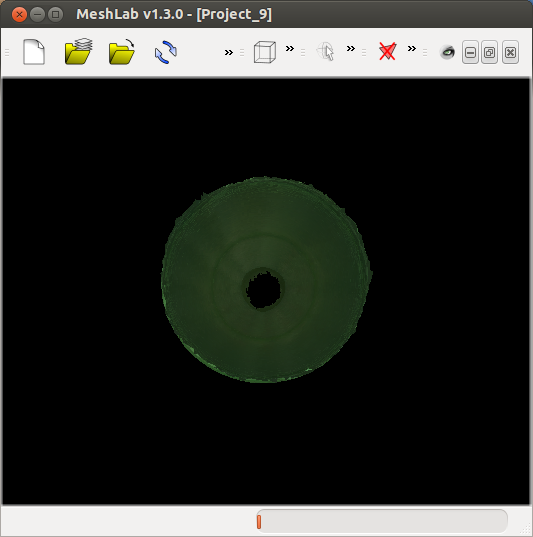}	\label{fig:Weights_831_20}	}
				\vspace*{-2mm}
			\caption{	Qualitative evaluation of the weight $\gamma_t$ of the energy function \eqref{eq:obj_regMatch}. 
					The images show the reconstruction of the objects \emph{water-bottle} and \emph{bowling-pin} (bottom view) for the weights $\gamma_t$: $0$, $1$, $5$, $10$, $15$ and $20$ (from left to right). 
			}
			\vspace*{-1mm}
			\label{fig:WeightsQualittive}
		\end{figure*}

\subsubsection{Qualitative evaluation}\label{sec:Experiments_RealisticData_Qualitative}

	\newcommand{\ImgSquizDetectRasterH} {-05mm}
	\newcommand{\ImgSquizDetectRasterHG}{-01.5mm}
	\newcommand{\ImgSquizDetectRasterV} {-07.5mm}
	\newcommand{\ImgSizeMultDetectRaster}{0.125}
	\begin{figure}[t]
		\vspace*{-2.0mm}
		\captionsetup[subfigure]{labelformat=empty}
		\centering
			\subfloat[subfigure BaselineDetector_CorrRaster2d_results 826 ff][]{	\includegraphics[trim=30mm 12mm 30mm 30mm, clip=true, height=\ImgSizeMultDetectRaster \textwidth]{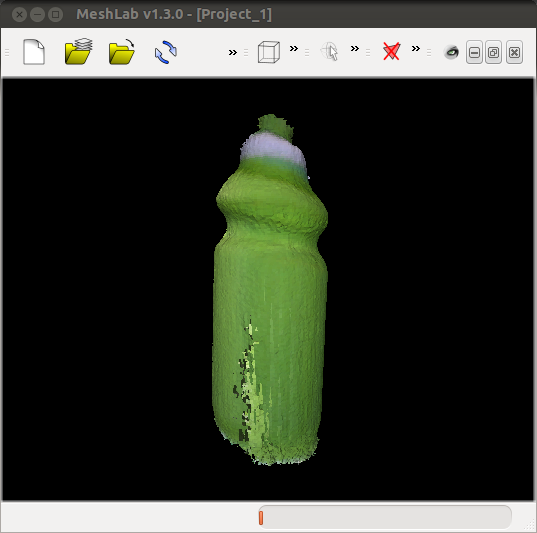}	\label{fig:BaselineDetector_CorrRaster2d_results_826_front}	}		\hspace*{\ImgSquizDetectRasterH}
			\subfloat[subfigure BaselineDetector_CorrRaster2d_results 826 bb][]{	\includegraphics[trim=30mm 12mm 30mm 30mm, clip=true, height=\ImgSizeMultDetectRaster \textwidth]{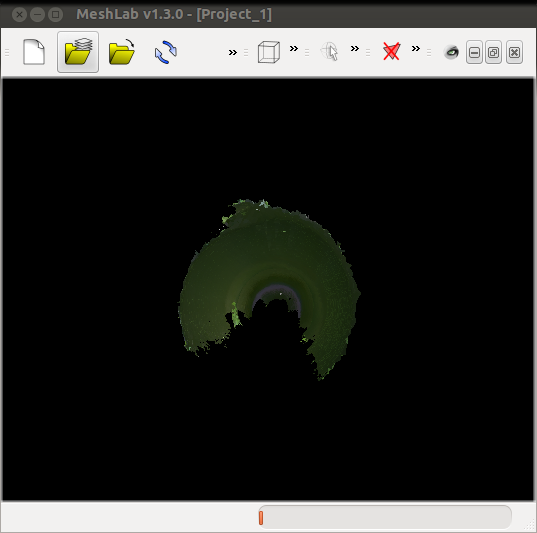}	\label{fig:BaselineDetector_CorrRaster2d_results_826_bottom}	}	  	\hspace*{\ImgSquizDetectRasterHG}
			\subfloat[subfigure BaselineDetector_CorrRaster2d_results 831 ff][]{	\includegraphics[trim=30mm 12mm 30mm 30mm, clip=true, height=\ImgSizeMultDetectRaster \textwidth]{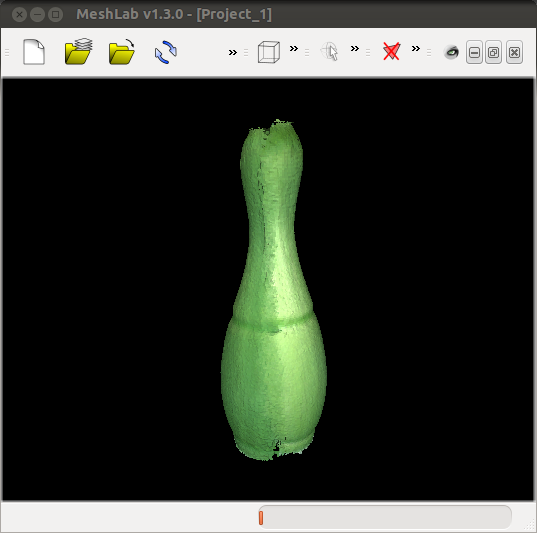}	\label{fig:BaselineDetector_CorrRaster2d_results_831_front}	}		\hspace*{\ImgSquizDetectRasterH}
			\subfloat[subfigure BaselineDetector_CorrRaster2d_results 831 bb][]{	\includegraphics[trim=30mm 12mm 30mm 30mm, clip=true, height=\ImgSizeMultDetectRaster \textwidth]{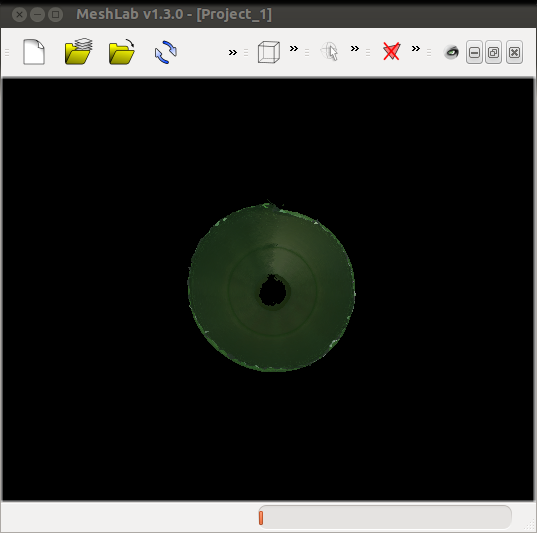}	\label{fig:BaselineDetector_CorrRaster2d_results_831_bottom}	}	\\	\vspace*{\ImgSquizDetectRasterV}
			\subfloat[subfigure BaselineDetector_CorrRaster2d_results 849 ff][]{	\includegraphics[trim=30mm 12mm 30mm 30mm, clip=true, height=\ImgSizeMultDetectRaster \textwidth]{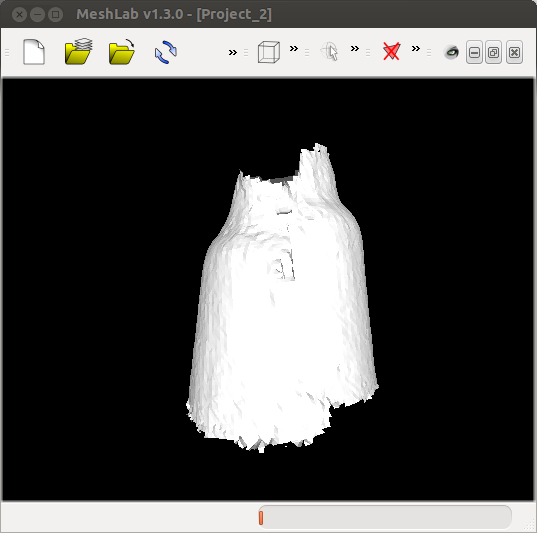}	\label{fig:BaselineDetector_CorrRaster2d_results_849_front}	}		\hspace*{\ImgSquizDetectRasterH}
			\subfloat[subfigure BaselineDetector_CorrRaster2d_results 849 bb][]{	\includegraphics[trim=30mm 12mm 30mm 30mm, clip=true, height=\ImgSizeMultDetectRaster \textwidth]{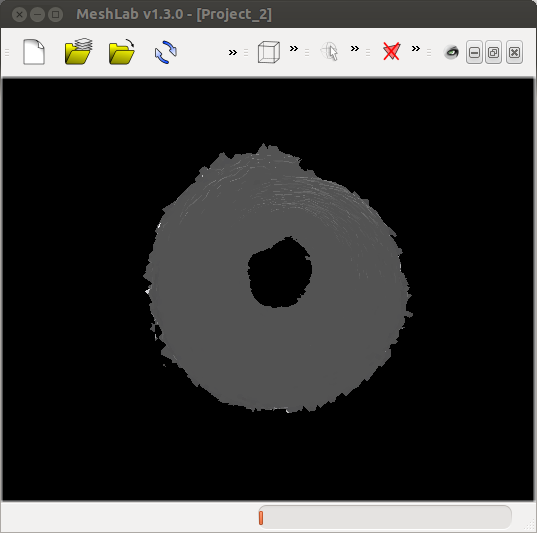}	\label{fig:BaselineDetector_CorrRaster2d_results_849_bottom}	}	  	\hspace*{\ImgSquizDetectRasterHG}
			\subfloat[subfigure BaselineDetector_CorrRaster2d_results 857 ff][]{	\includegraphics[trim=30mm 12mm 30mm 30mm, clip=true, height=\ImgSizeMultDetectRaster \textwidth]{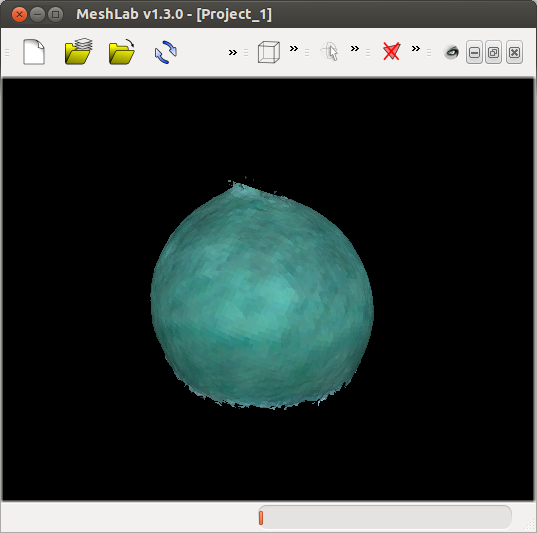}	\label{fig:BaselineDetector_CorrRaster2d_results_857_front}	}		\hspace*{\ImgSquizDetectRasterH}
			\subfloat[subfigure BaselineDetector_CorrRaster2d_results 857 bb][]{	\includegraphics[trim=30mm 12mm 30mm 30mm, clip=true, height=\ImgSizeMultDetectRaster \textwidth]{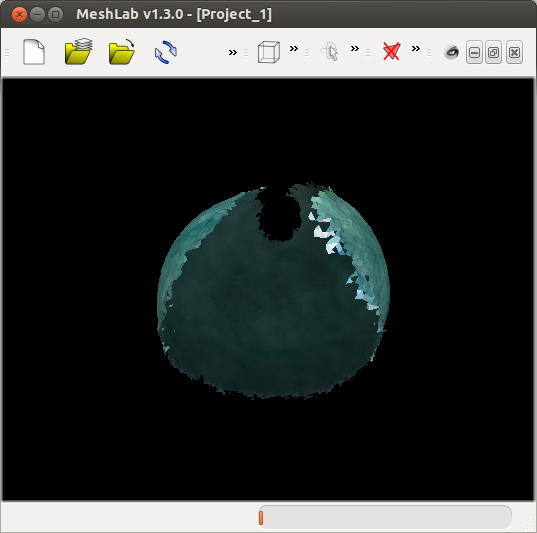}	\label{fig:BaselineDetector_CorrRaster2d_results_857_bottom}	}	
			\vspace*{-2mm}
		\caption{
				\reviewChange
				{
				When a 
				\emph{contact detector} is used instead of the contact points based on a \emph{hand tracker}, the reconstruction fails.   
				For each object the front and the bottom view are shown. 
				}
		}
		\label{fig:BaselineDetector_CorrRaster2d_results}
		\vspace*{-1.0mm}
	\end{figure}

		\newcommand{\ImgSquizNoIcpH}{-06mm}
		\newcommand{\ImgSquizNoIcpV}{-08mm}
		\newcommand{\ImgSizeMultNoIcp}{0.14}	
		\begin{figure}[t]
		\captionsetup[subfigure]{labelformat=empty}
		\centering
			\subfloat[subfigure BowlingPin_NoIcp 1][]{	\includegraphics[trim=05mm 15mm 05mm 30mm, clip=true, height=\ImgSizeMultNoIcp  \textwidth]{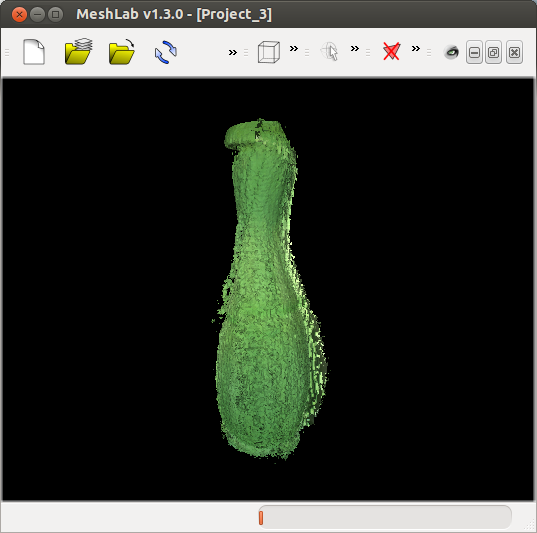}	\label{fig:BowlingPin_NoIcp_tsdf}	}		\hspace*{\ImgSquizNoIcpH}
			\subfloat[subfigure BowlingPin_NoIcp 2][]{	\includegraphics[trim=05mm 15mm 05mm 30mm, clip=true, height=\ImgSizeMultNoIcp  \textwidth]{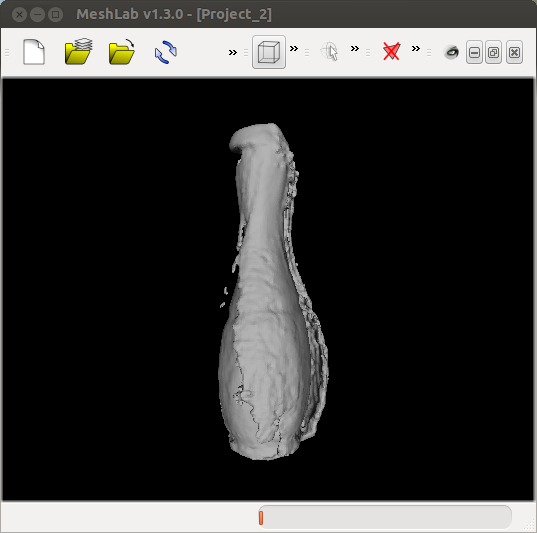}	\label{fig:BowlingPin_NoIcp_poisson}	}	
			\vspace*{-2mm}
			\caption{	
					\reviewChange
					{
					Qualitative results for the \emph{bowling-pin} object without the term $E_{icp}$ of Equation \eqref{eq:obj_reg_icp}. In this case, the point clouds are not well aligned. 
					}
			}
		\label{fig:BowlingPin_NO_ICP}
		\end{figure}

	\begin{table}[t]
			\vspace*{+1.5mm}
		\footnotesize 
		\begin{center}
			\caption{	
					\reviewChange
					{
					Quantitative evaluation of the pairwise registration based on annotated pairs of frames. 
					We assess the performance of both 
					the proposed pipeline based on hand pose $E_{contact}$ as described in Equation \eqref{eq:obj_reg_contact}, 
					as well as the detector-based baseline $E_{detect}$ as described in Equation  \eqref{eq:obj_reg_detector}. 
					We report the mean and the standard deviation over all the sampled frame pairs of all sequences in millimeters. 
					}
			}
			\label{fig:QResultsDETECTOR}
			\setlength{\tabcolsep}{1pt}	
			\begin{tabular}{|l|c|c|c|c|c|c|c|}																																				\hhline{-~--~~}
			
				\multicolumn{1}{|c|}{~~Energy~~} 	& 
				\multicolumn{1}{ c|}{} 					& 
				~mean~							& 
				~st.dev.~						& 
				\multicolumn{1}{c}{}					& 
				\multicolumn{1}{c}{} 																																				\\	\hhline{-~--~~}
				\noalign{\smallskip}																																					\hhline{-~--~-}
														{$E_{contact}$+$E_{visual}$}	& {}	& { $1.67$ }	& {~$0.95$~} 	& {} & \multirow{4}{*}{\centering\begin{turn}{90}mm\end{turn}}		\\	\hhline{-~--~~}
							{$E_{contact}$}	& {}	& {~$1.64$~}	& {~$0.88$~} 	& {} & {} 								\\	\hhline{-~--~~}
							{$E_{detector}$+$E_{visual}$}	& {}	& { $1.73$ }	& { $1.08$ } 	& {} & {} 								\\	\hhline{-~--~~}
							{$E_{detector}$}	& {}	& { $1.80$ }	& {~$1.12$~} 	& {} & {} 								\\	\hhline{-~--~-}
			\end{tabular}
		\end{center}
		\vspace*{-2mm}
	\end{table}

	Although the quantitative evaluation is informative, a qualitative evaluation can give further intuition about the effectiveness of the system and the influence of its parameters. 
	We therefore show in Figure \ref{fig:WeightsQualittive} the mesh extracted from the TSDF volume of our pipeline for a number of different values for the weight $\gamma_t$. 
	The experiment is done for the two objects where the influence of $\gamma_t$ can be easily observed visually. 
	As expected, the reconstruction without the use of hand motion capture results in a degenerate alignment. 
	The incorporation of \emph{contact correspondences} immediately improves the reconstruction, driving the alignment process according to the spatiotemporal movement of the contact fingers. 
	A low value, however, leads only to a partial reconstruction. 
	A sensible choice seems to be a value between $10$ and $30$, while for bigger values some small alignment artifacts appear. 
	For our experiments we choose $\gamma_t=15$. 
	
	\reviewChange
	{
	While Figure \ref{fig:BaselineDetector_CorrRaster2d_results} shows the reconstruction when the hand tracker is replaced by a detector, 
	Figure  \ref{fig:BowlingPin_NO_ICP} shows the reconstruction of the \emph{bowling-pin} when ICP, as described in Equation \eqref{eq:obj_reg_icp}, is not used. 
	In both cases, the point clouds are not well aligned.  
	}

	Figure \ref{fig:My_Results_VS_competition} shows the best reconstruction of three runs by \emph{KinFu} and \emph{Skanect} in comparison to our pipeline, 
	both with and without the use of hands and hand motion data. 
	The images show that the reconstruction without hands is similar across different systems and results in a degenerate 3d representation of the object. 
	The incorporation of hand motion capture in the reconstruction plays clearly a vital role, leading to the effective reconstruction of the full surface of the object.

	Although Figure \ref{fig:My_Results_VS_competition} compares the TSDF meshes, more detailed results are shown in Figure \ref{fig:My_Results}. 
	The camera poses are reconstructed effectively, showing not only the rotational movement during the scanning process, but also the type and intensity of hand-object interaction. 
	The water-tight meshes that are shown compose the final output of our system. 
	The resulting reconstruction renders our approach the first in-hand scanning system to cope with the reconstruction of symmetric objects, while also showing prospects of future practical applications.

		\newcommand{\ImgSquizCompetitionH}{-05mm}
		\newcommand{\ImgSquizCompetitionV}{-08mm}
		\newcommand{\ImgSizeMultCompetition}{0.10}
		\begin{figure}[t]
		\vspace*{-3mm}
		\captionsetup[subfigure]{labelformat=empty}
		\centering
			\subfloat[subfigure competition 826 1][]{	\includegraphics[trim=20mm 10mm 30mm 23mm, clip=true, width=\ImgSizeMultCompetition \textwidth]{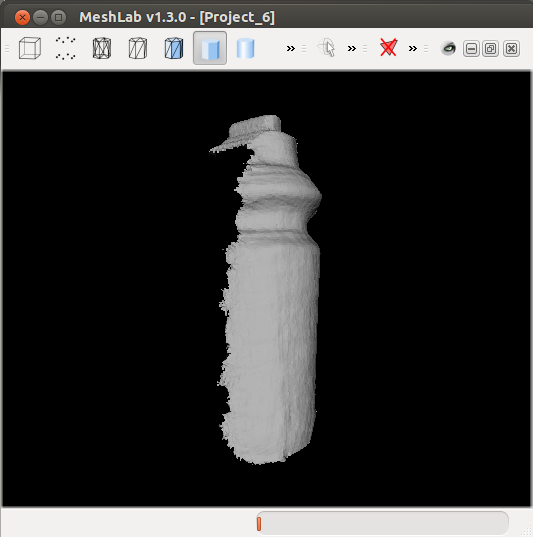}	\label{fig:OursVsCompet_826_1}	}		\hspace*{\ImgSquizCompetitionH}
			\subfloat[subfigure competition 826 2][]{	\includegraphics[trim=20mm 10mm 30mm 23mm, clip=true, width=\ImgSizeMultCompetition \textwidth]{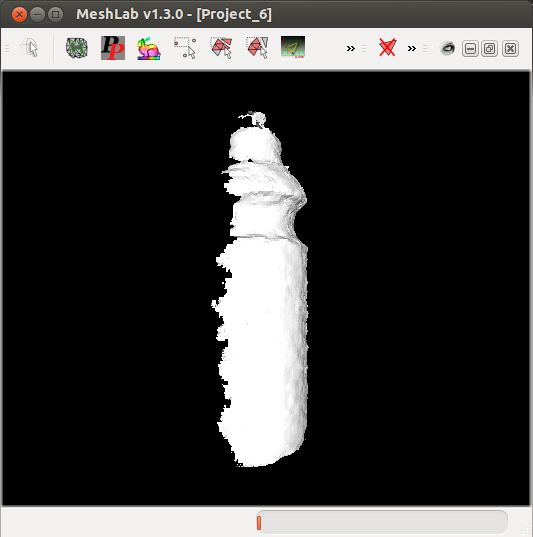}	\label{fig:OursVsCompet_826_2}	}		\hspace*{\ImgSquizCompetitionH}
			\subfloat[subfigure competition 826 4][]{	\includegraphics[trim=20mm 10mm 30mm 23mm, clip=true, width=\ImgSizeMultCompetition \textwidth]{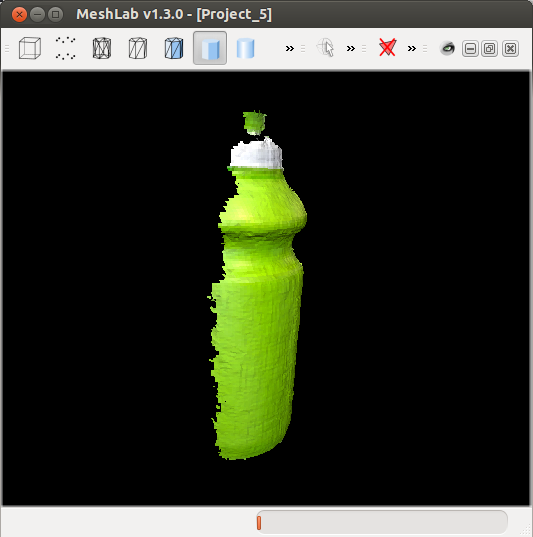}	\label{fig:OursVsCompet_826_4}	}		\hspace*{\ImgSquizCompetitionH}
			\subfloat[subfigure competition 826 5][]{	\includegraphics[trim=20mm 10mm 30mm 23mm, clip=true, width=\ImgSizeMultCompetition \textwidth]{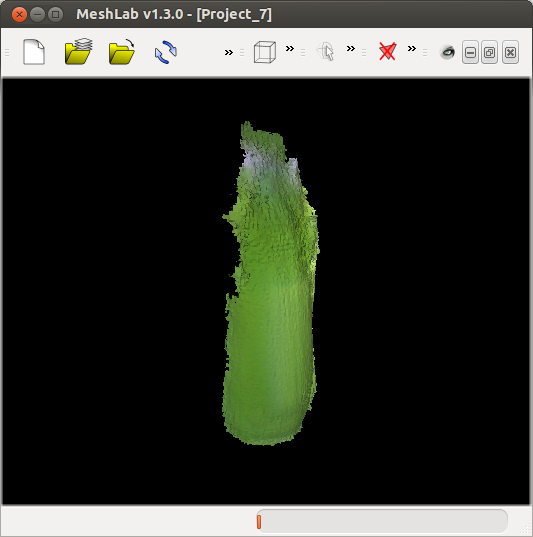}	\label{fig:OursVsCompet_826_5}	}		\hspace*{\ImgSquizCompetitionH}
			\subfloat[subfigure competition 826 6][]{	\includegraphics[trim=20mm 10mm 30mm 23mm, clip=true, width=\ImgSizeMultCompetition \textwidth]{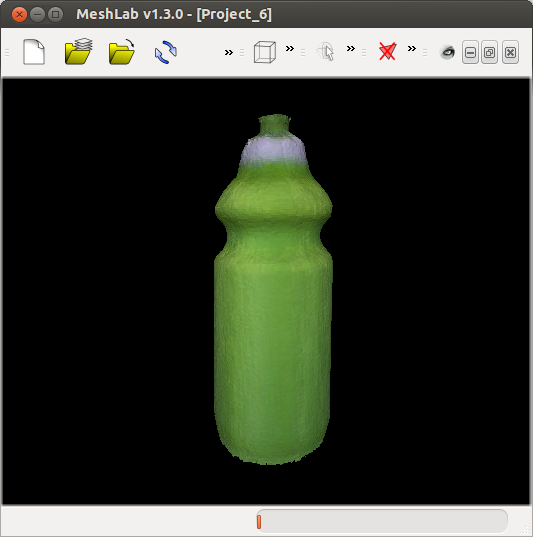}	\label{fig:OursVsCompet_826_6}	}	\\	\vspace*{\ImgSquizCompetitionV}			
			
			\subfloat[subfigure competition 831 1][]{	\includegraphics[trim=20mm 10mm 30mm 23mm, clip=true, width=\ImgSizeMultCompetition \textwidth]{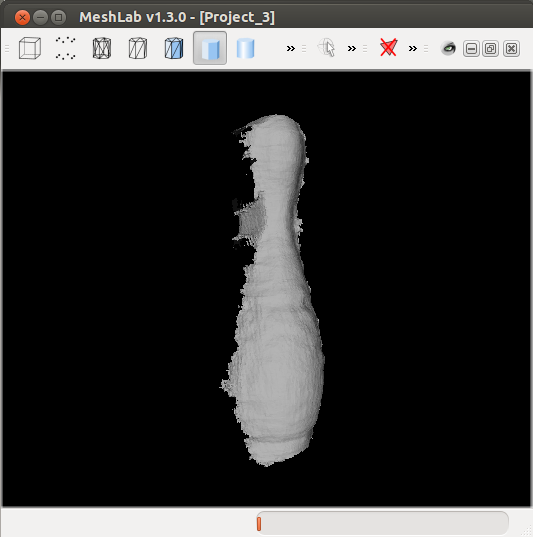}	\label{fig:OursVsCompet_831_1}	}		\hspace*{\ImgSquizCompetitionH}
			\subfloat[subfigure competition 831 2][]{	\includegraphics[trim=20mm 10mm 30mm 23mm, clip=true, width=\ImgSizeMultCompetition \textwidth]{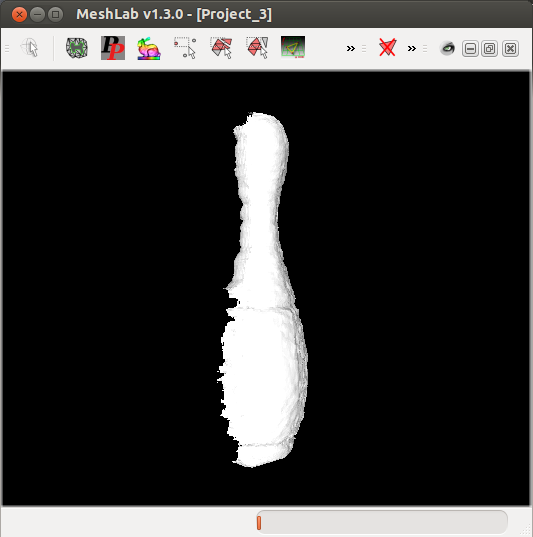}	\label{fig:OursVsCompet_831_2}	}		\hspace*{\ImgSquizCompetitionH}
			\subfloat[subfigure competition 831 4][]{	\includegraphics[trim=20mm 10mm 30mm 23mm, clip=true, width=\ImgSizeMultCompetition \textwidth]{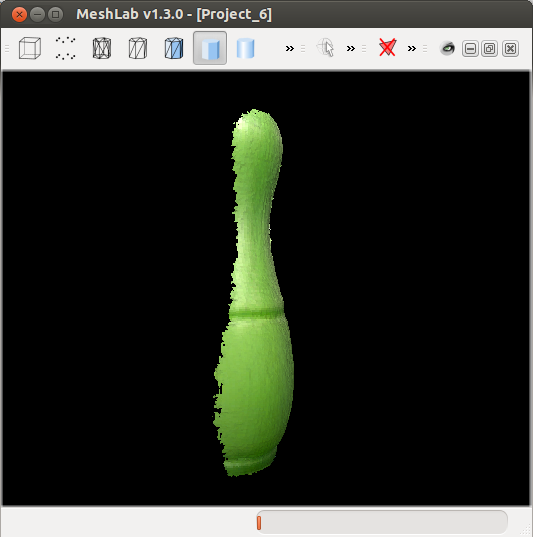}	\label{fig:OursVsCompet_831_4}	}		\hspace*{\ImgSquizCompetitionH}
			\subfloat[subfigure competition 831 5][]{	\includegraphics[trim=20mm 10mm 30mm 23mm, clip=true, width=\ImgSizeMultCompetition \textwidth]{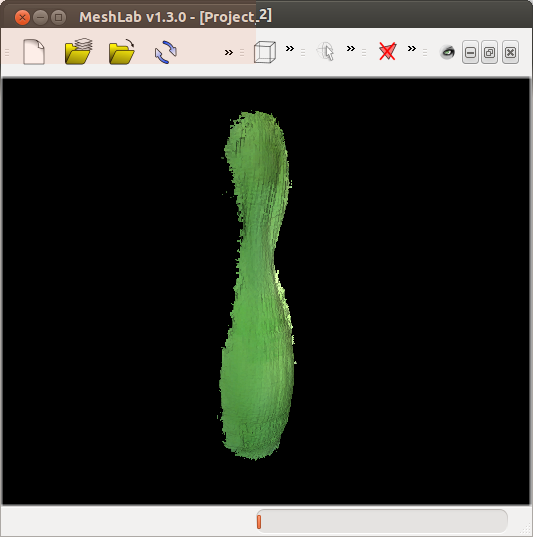}	\label{fig:OursVsCompet_831_5}	}		\hspace*{\ImgSquizCompetitionH}
			\subfloat[subfigure competition 831 6][]{	\includegraphics[trim=20mm 10mm 30mm 23mm, clip=true, width=\ImgSizeMultCompetition \textwidth]{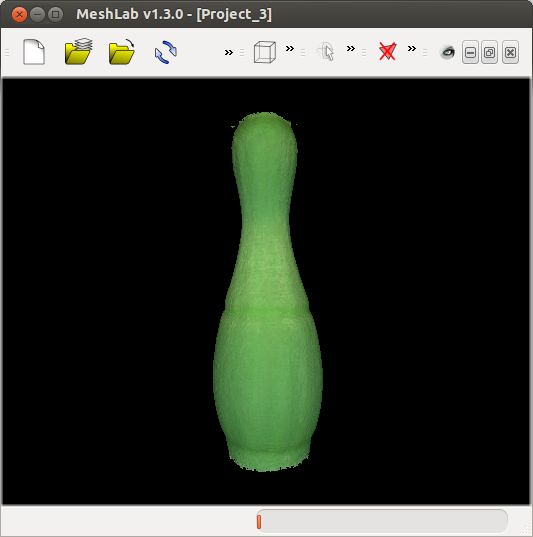}	\label{fig:OursVsCompet_831_6}	}	\\	\vspace*{\ImgSquizCompetitionV}	
			
			\subfloat[subfigure competition 849 1][]{	\includegraphics[trim=20mm 10mm 30mm 23mm, clip=true, width=\ImgSizeMultCompetition \textwidth]{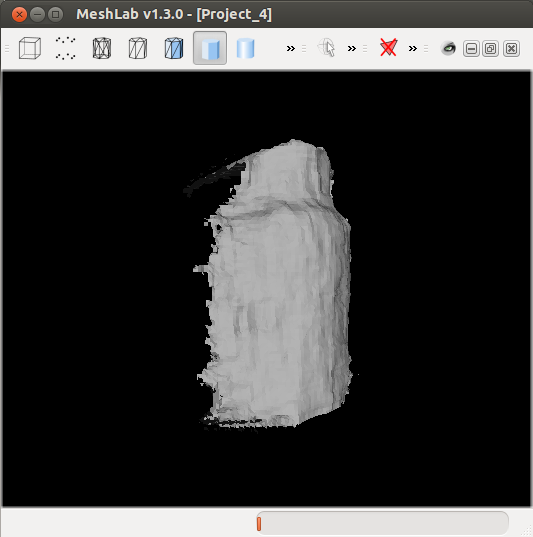}	\label{fig:OursVsCompet_849_1}	}		\hspace*{\ImgSquizCompetitionH}
			\subfloat[subfigure competition 849 2][]{	\includegraphics[trim=20mm 10mm 30mm 23mm, clip=true, width=\ImgSizeMultCompetition \textwidth]{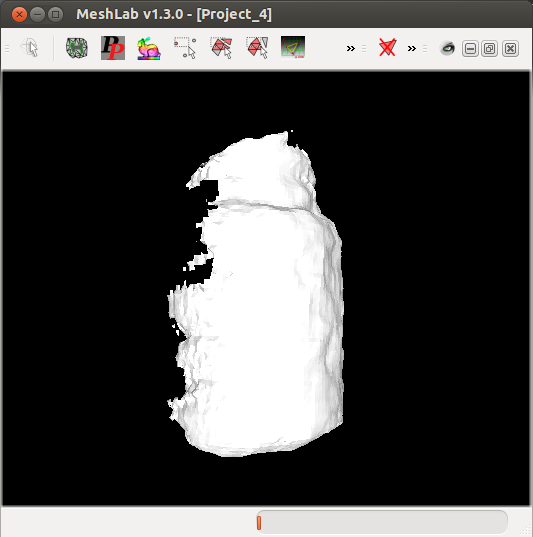}	\label{fig:OursVsCompet_849_2}	}		\hspace*{\ImgSquizCompetitionH}
			\subfloat[subfigure competition 849 4][]{	\includegraphics[trim=20mm 10mm 30mm 23mm, clip=true, width=\ImgSizeMultCompetition \textwidth]{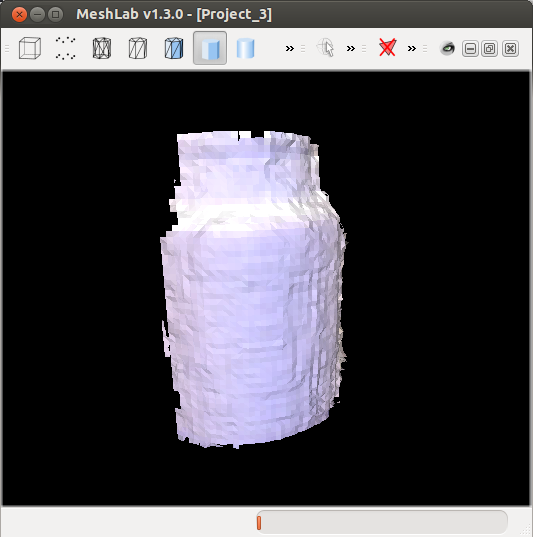}	\label{fig:OursVsCompet_849_4}	}		\hspace*{\ImgSquizCompetitionH}
			\subfloat[subfigure competition 849 5][]{	\includegraphics[trim=20mm 10mm 30mm 23mm, clip=true, width=\ImgSizeMultCompetition \textwidth]{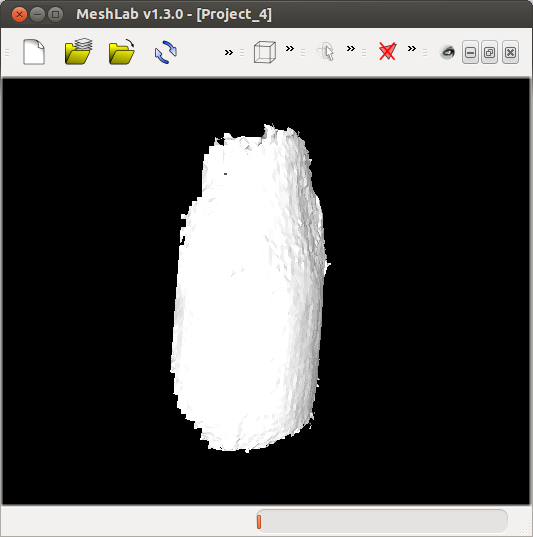}	\label{fig:OursVsCompet_849_5}	}		\hspace*{\ImgSquizCompetitionH}
			\subfloat[subfigure competition 849 6][]{	\includegraphics[trim=20mm 10mm 30mm 23mm, clip=true, width=\ImgSizeMultCompetition \textwidth]{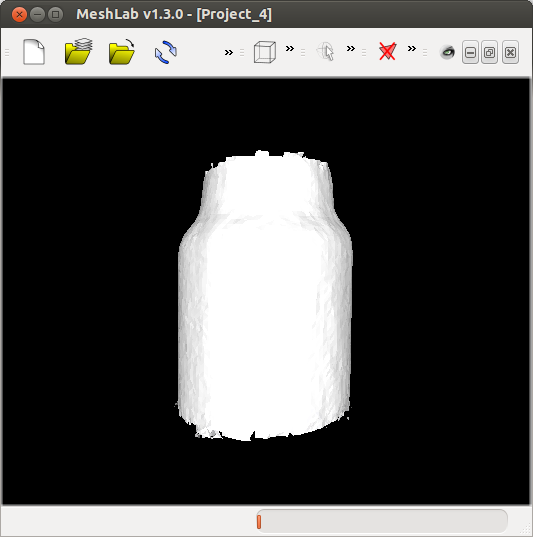}	\label{fig:OursVsCompet_849_6}	}	\\	\vspace*{\ImgSquizCompetitionV}
			
			\subfloat[subfigure competition 857 1][]{	\includegraphics[trim=20mm 10mm 30mm 23mm, clip=true, width=\ImgSizeMultCompetition \textwidth]{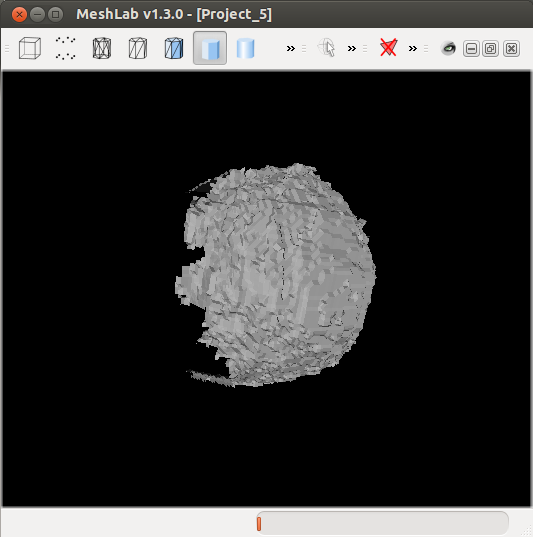}	\label{fig:OursVsCompet_857_1}	}		\hspace*{\ImgSquizCompetitionH}
			\subfloat[subfigure competition 857 2][]{	\includegraphics[trim=20mm 10mm 30mm 23mm, clip=true, width=\ImgSizeMultCompetition \textwidth]{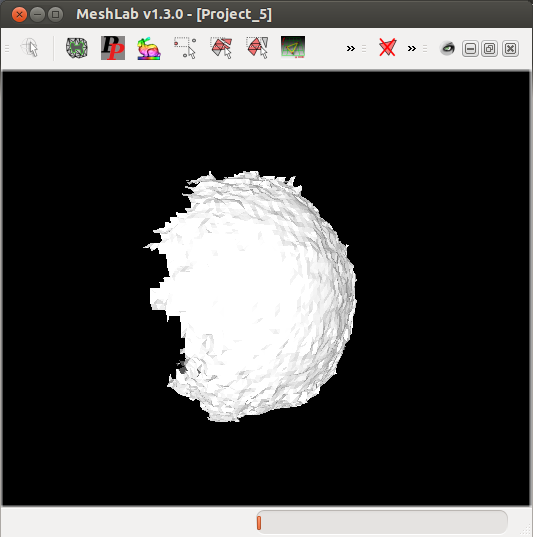}	\label{fig:OursVsCompet_857_2}	}		\hspace*{\ImgSquizCompetitionH}
			\subfloat[subfigure competition 857 4][]{	\includegraphics[trim=20mm 10mm 30mm 23mm, clip=true, width=\ImgSizeMultCompetition \textwidth]{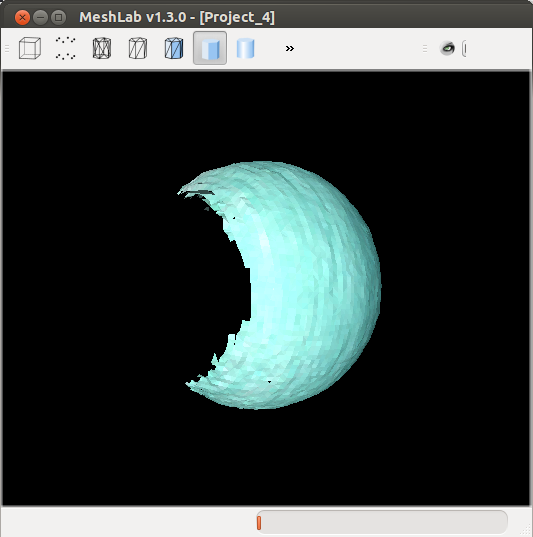}	\label{fig:OursVsCompet_857_4}	}		\hspace*{\ImgSquizCompetitionH}
			\subfloat[subfigure competition 857 5][]{	\includegraphics[trim=20mm 10mm 30mm 23mm, clip=true, width=\ImgSizeMultCompetition \textwidth]{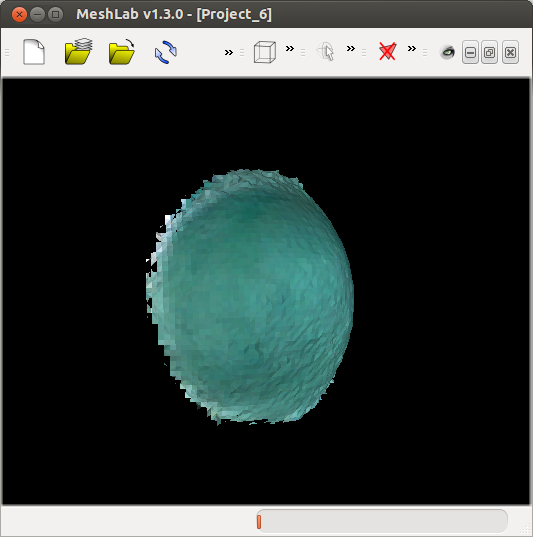}	\label{fig:OursVsCompet_857_5}	}		\hspace*{\ImgSquizCompetitionH}
			\subfloat[subfigure competition 857 6][]{	\includegraphics[trim=20mm 10mm 30mm 23mm, clip=true, width=\ImgSizeMultCompetition \textwidth]{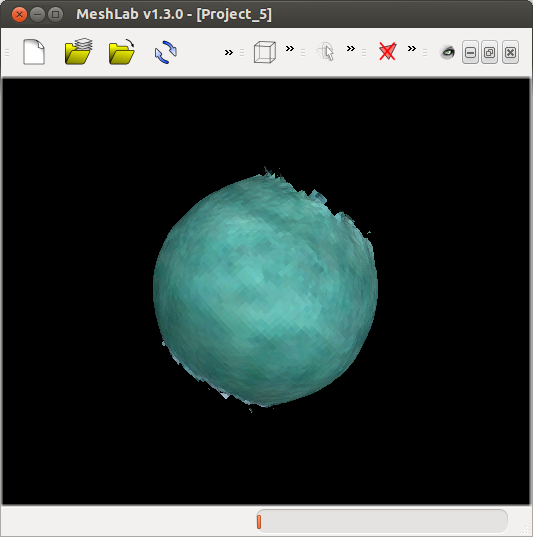}	\label{fig:OursVsCompet_857_6}	}	
			\vspace*{-2mm}
		\caption{	Qualitative comparison of different in-hand scanning systems for all four objects of Figure \ref{fig:photo_objectsAll}. 
				We visualize the meshes extracted from the TSDF volume. 
				From left to right, each row contains the result of: 
				(a) KinFu, 
				(b) Skanect, 
				(c) Our pipeline with a turntable and without hand motion data, 
				(d) Our pipeline with in-hand scanning but without hand motion data, 
				(e) Our pipeline with in-hand scanning that includes hand motion data (the proposed setup). 
				Only the combination of in-hand scanning with hand motion data succeeds in reconstructing all symmetric objects. 
		}
		\label{fig:My_Results_VS_competition}
		\end{figure}

		\newcommand{\ImgSquizMyresultsH}{-05mm}
		\newcommand{\ImgSquizMyresultsV}{-08.0mm}
		\newcommand{\ImgSizeMultMyresults}{0.120}
		\begin{figure}[t]
		\vspace*{-3.0mm}
		\captionsetup[subfigure]{labelformat=empty}
		\centering
			\subfloat[subfigure MyResults 826 1][]{	\includegraphics[trim=10mm 10mm 00mm 30mm, clip=true, height=\ImgSizeMultMyresults \textwidth]{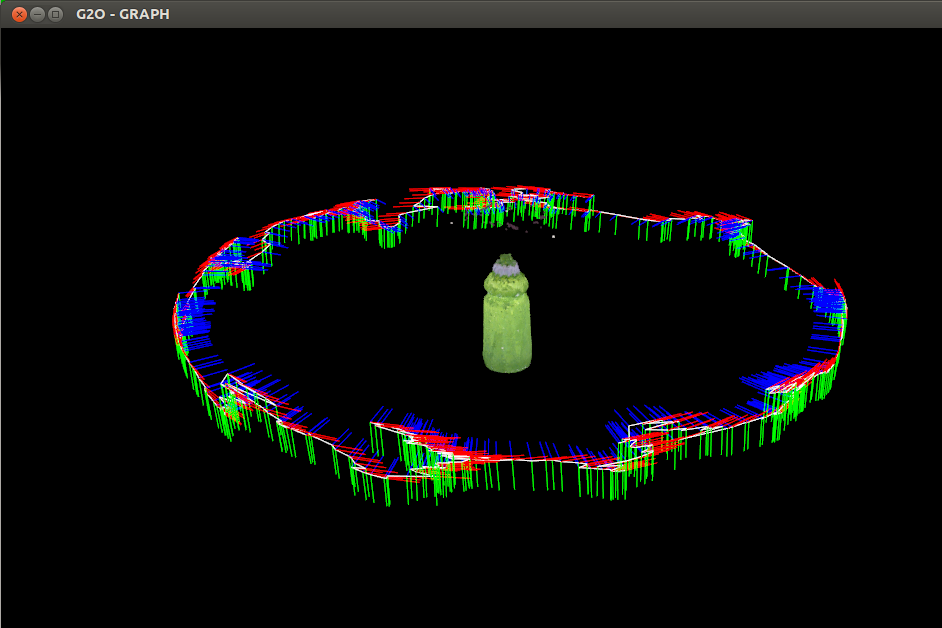}	\label{fig:MyResults_826_1}	}		\hspace*{\ImgSquizMyresultsH}
			\subfloat[subfigure MyResults 826 2][]{	\includegraphics[trim=20mm 20mm 15mm 30mm, clip=true, height=\ImgSizeMultMyresults \textwidth]{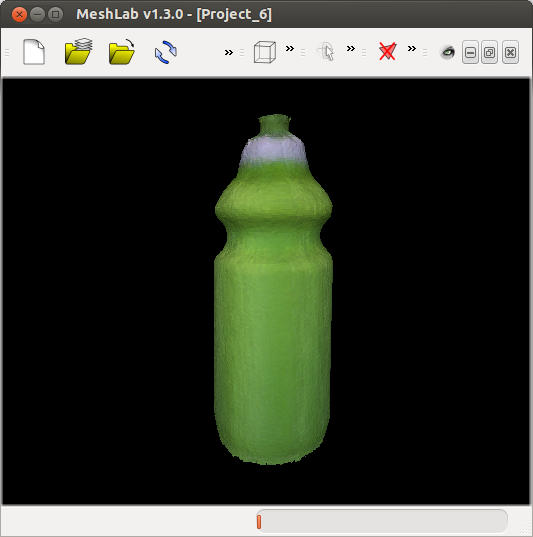}	\label{fig:MyResults_826_2}	}		\hspace*{\ImgSquizMyresultsH}
			\subfloat[subfigure MyResults 826 3][]{	\includegraphics[trim=20mm 20mm 15mm 30mm, clip=true, height=\ImgSizeMultMyresults \textwidth]{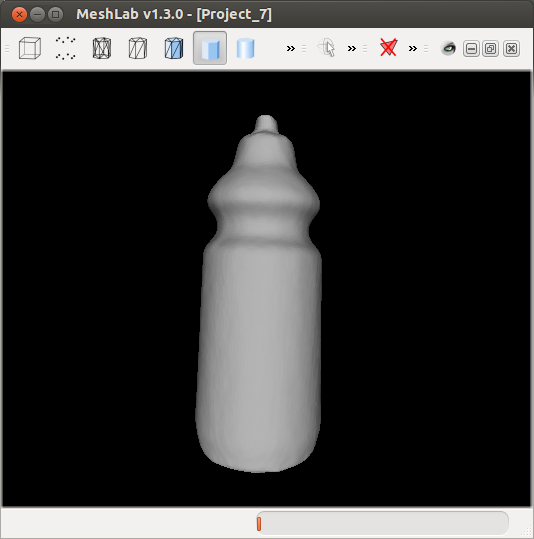}	\label{fig:MyResults_826_3}	}	\\	\vspace*{\ImgSquizMyresultsV}			
			
			\subfloat[subfigure MyResults 831 1][]{	\includegraphics[trim=10mm 10mm 00mm 30mm, clip=true, height=\ImgSizeMultMyresults \textwidth]{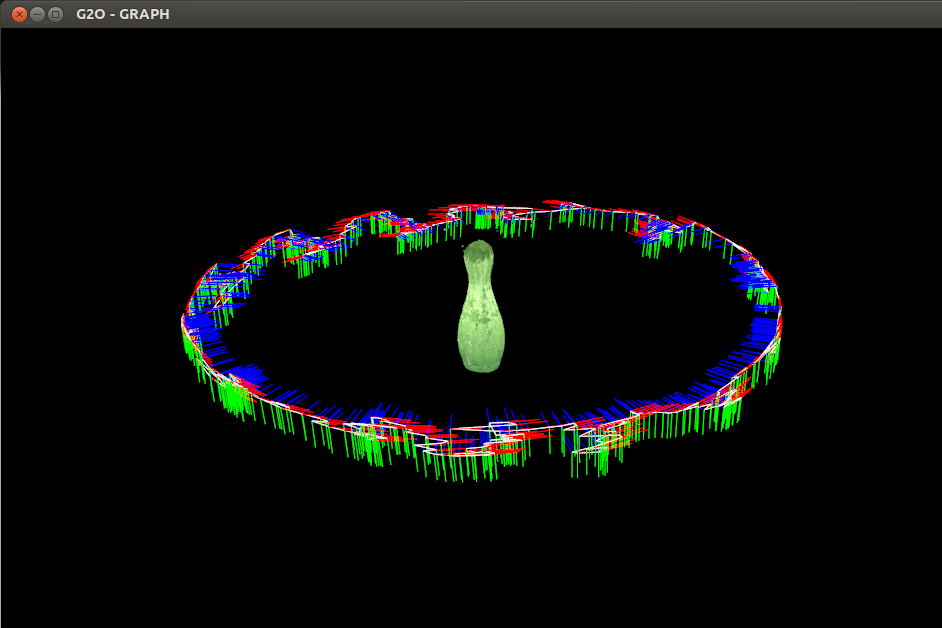}	\label{fig:MyResults_831_1}	}		\hspace*{\ImgSquizMyresultsH}
			\subfloat[subfigure MyResults 831 2][]{	\includegraphics[trim=20mm 20mm 15mm 30mm, clip=true, height=\ImgSizeMultMyresults \textwidth]{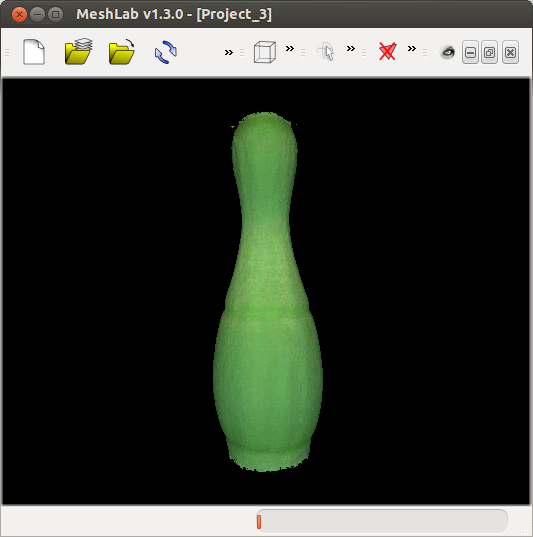}	\label{fig:MyResults_831_2}	}		\hspace*{\ImgSquizMyresultsH}
			\subfloat[subfigure MyResults 831 3][]{	\includegraphics[trim=20mm 20mm 15mm 30mm, clip=true, height=\ImgSizeMultMyresults \textwidth]{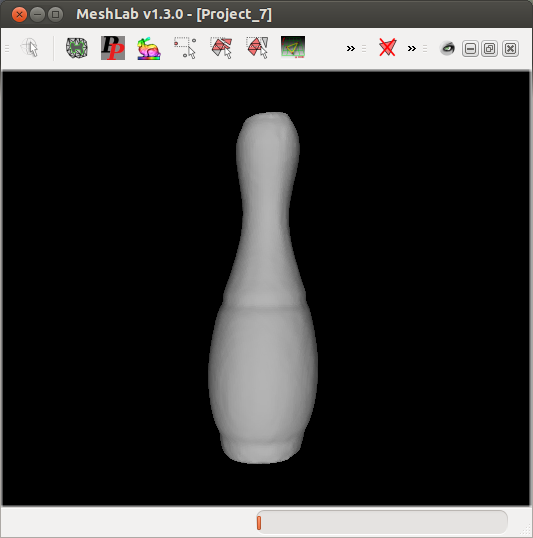}	\label{fig:MyResults_831_3}	}	\\	\vspace*{\ImgSquizMyresultsV}
			
			\subfloat[subfigure MyResults 849 1][]{	\includegraphics[trim=10mm 10mm 00mm 30mm, clip=true, height=\ImgSizeMultMyresults \textwidth]{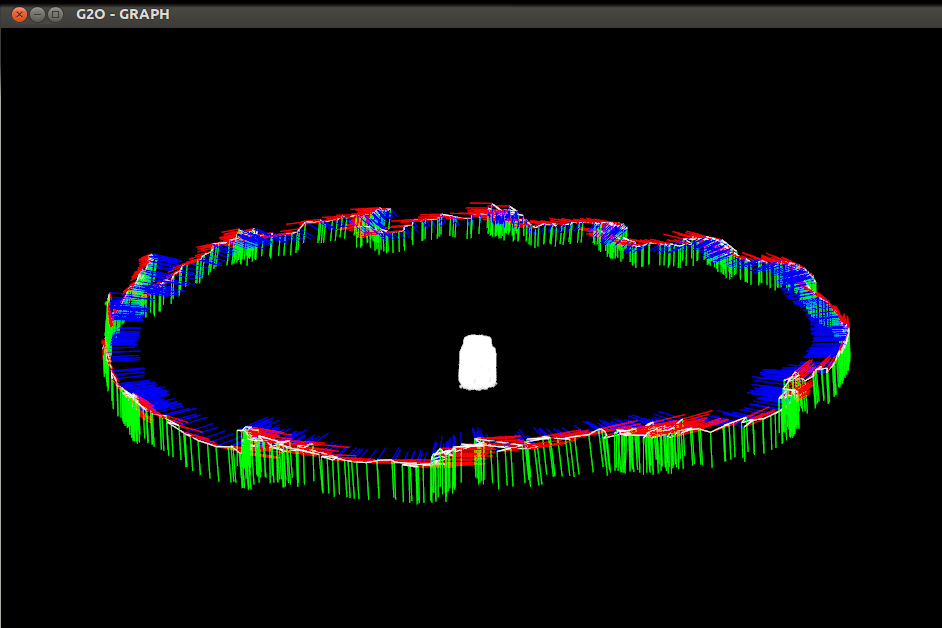}	\label{fig:MyResults_849_1}	}		\hspace*{\ImgSquizMyresultsH}
			\subfloat[subfigure MyResults 849 2][]{	\includegraphics[trim=20mm 20mm 15mm 30mm, clip=true, height=\ImgSizeMultMyresults \textwidth]{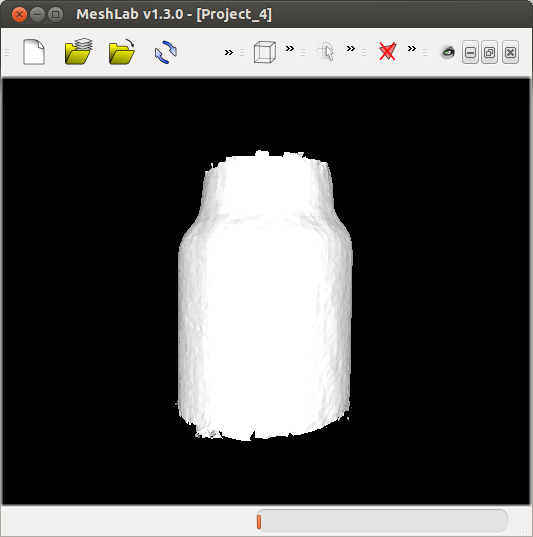}	\label{fig:MyResults_849_2}	}		\hspace*{\ImgSquizMyresultsH}
			\subfloat[subfigure MyResults 849 3][]{	\includegraphics[trim=20mm 20mm 15mm 30mm, clip=true, height=\ImgSizeMultMyresults \textwidth]{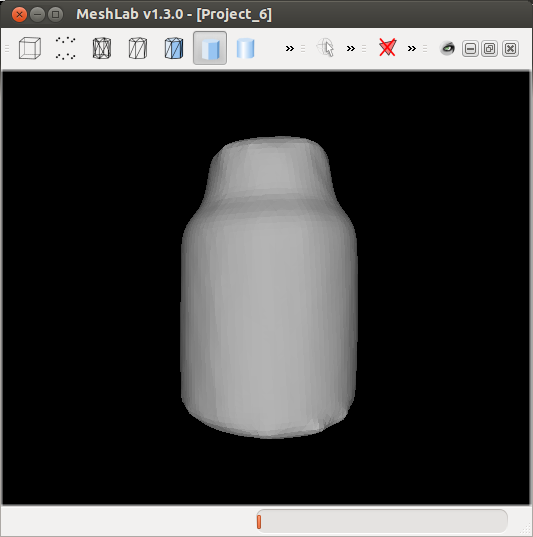}	\label{fig:MyResults_849_3}	}	\\	\vspace*{\ImgSquizMyresultsV}			
			
			\subfloat[subfigure MyResults 857 1][]{	\includegraphics[trim=10mm 10mm 00mm 30mm, clip=true, height=\ImgSizeMultMyresults \textwidth]{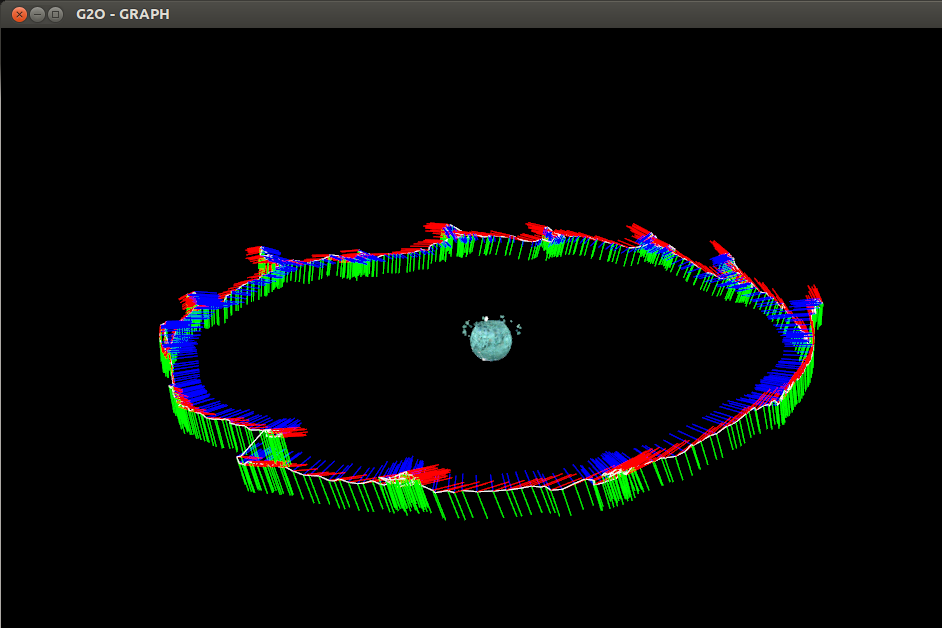}	\label{fig:MyResults_857_1}	}		\hspace*{\ImgSquizMyresultsH}
			\subfloat[subfigure MyResults 857 2][]{	\includegraphics[trim=20mm 20mm 15mm 30mm, clip=true, height=\ImgSizeMultMyresults \textwidth]{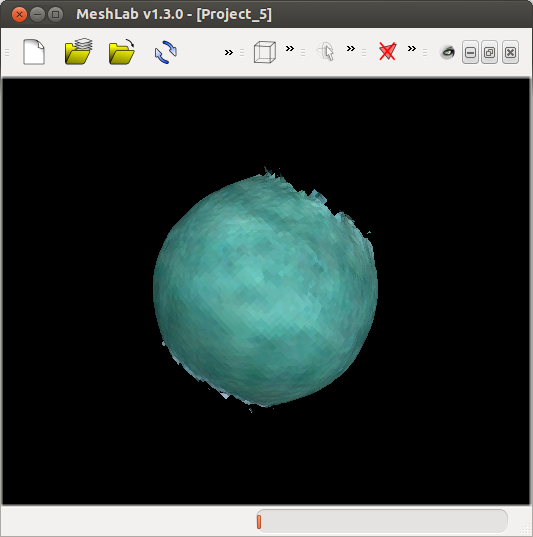}	\label{fig:MyResults_857_2}	}		\hspace*{\ImgSquizMyresultsH}
			\subfloat[subfigure MyResults 857 3][]{	\includegraphics[trim=20mm 20mm 15mm 30mm, clip=true, height=\ImgSizeMultMyresults \textwidth]{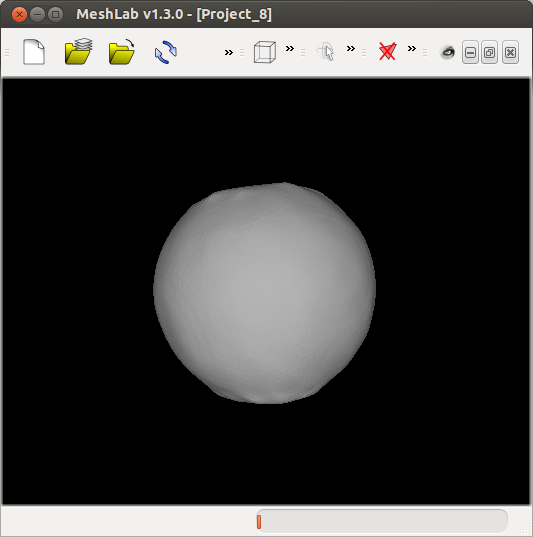}	\label{fig:MyResults_857_3}	}	
			\vspace*{-2mm}
		\caption{	Qualitative results of our pipeline for all four objects of Figure \ref{fig:photo_objectsAll} when a hand rotates the object in front of the camera. 
				The left images show the reconstructed camera poses. 
				The poses follow a circular path, whose shape signifies the type of hand-object interaction during the rotation. 
				The middle images show the mesh that is acquired with marching cubes from the TSDF volume, while 
				the right ones show the final water-tight mesh that is acquired with Poisson reconstruction. 
		}
		\label{fig:My_Results}
		\vspace*{+0.8mm}
		\end{figure}


\reviewChange
{
\subsection{Limitations}\label{sec:limitation_FutureWork}
While our approach does not depend on a specific hand tracker, it only works if the hand tracker does not fail. 
At the moment only end-effectors are considered, therefore cases where there is only contact with the palm are not handled, but the approach can be extended to more general contact points. 
Moreover the case of fingers slipping over the manipulated object is not handled currently, but this could be addressed by using a hand tracker that estimates forces~\cite{Pham_2015_CVPR}. 
}


\vspace*{-0.5mm}
\section{Conclusion}\label{sec:Conclusion}

	While existing in-hand scanning systems discard information originating from the hand, we have proposed an approach that successfully incorporates the 3d motion information of the manipulating hand for 3d object reconstruction. 
	In that respect, the visual correspondences based on geometric and texture features are combined with contact correspondences that are inferred from the manipulating hand. 
	In our quantitative and qualitative experiments we show that our approach successfully reconstructs the 3d shape of four highly symmetric and textureless objects.


\section{Acknowledgments}\label{sec:Acknowledgments}

	We 
	gratefully 
	acknowledge 
	the help of Abhilash Srikantha with the Hough forest detector 	and Pablo Aponte with KinFu. 
	Financial support was provided by the DFG Emmy Noether program (GA 1927/1-1). 
	Dimitrios Tzionas would like to dedicate this work in the memory of his grandmother Olga Matoula. 
	\\


{\small
\bibliographystyle{ieee}
\bibliography{egbib}
}

\end{document}